\documentclass[10pt,twocolumn,letterpaper]{article}

\usepackage[pagenumbers]{wacv} 

\definecolor{wacvblue}{rgb}{0.21,0.49,0.74}
\usepackage[pagebackref,breaklinks,colorlinks,allcolors=wacvblue]{hyperref}
\usepackage[utf8]{inputenc} 
\usepackage[T1]{fontenc}    
\usepackage{hyperref}       
\usepackage{url}            
\usepackage{amsfonts}       
\usepackage{nicefrac}       
\usepackage{microtype}      
\usepackage[dvipsnames]{xcolor}        
\usepackage{amssymb}
\usepackage{mathtools}
\usepackage{amsthm}
\usepackage{pifont}
\usepackage{bbm}
\usepackage{mathtools}
\definecolor{ForestGreen}{RGB}{34, 139, 34}
\usepackage[normalem]{ulem}
\usepackage{comment}
\usepackage{booktabs}
\usepackage{multirow}
\usepackage{algorithmic}
\usepackage{algorithm}
\usepackage{subcaption}
\usepackage[accsupp]{axessibility}  

\newcommand{\cmark}{\ding{51}}%
\newcommand{\xmark}{\ding{55}}%
\usepackage{xcolor}

\def\wacvPaperID{274} 
\def\confName{WACV}
\def\confYear{2026}

\title{Empowering Source-Free Domain Adaptation via MLLM-Guided
Reliability-Based Curriculum Learning}

\author{
Dongjie Chen$^{1}$\thanks{Equal contribution. \\
Correspondence to: Dongjie Chen \texttt{$<$dongjiech@gmail.com$>$}.}\qquad
Kartik Patwari$^{1}$\footnotemark[1]\qquad
Zhengfeng Lai$^{1}$\qquad
Xiaoguang Zhu$^{1}$\\
Sen-ching Cheung$^{1,2}$\qquad
Chen-Nee Chuah$^{1}$\\[7pt]
$^{1}$ University of California, Davis \qquad
$^{2}$ University of Kentucky
}

\begin{document}

\maketitle

\begin{abstract}
    Existing SFDA methods struggle to fully use pre-trained knowledge and often rely on a single model’s predictions or handcrafted prompts, limiting robustness under domain shift. Multimodal Large Language Models (MLLMs) offer a promising alternative: they encode rich visual-semantic knowledge and generalize well without task-specific tuning. However, their use in SFDA is hindered by instruction-following failures, inconsistent outputs, and high inference costs. We propose Reliability-based Curriculum Learning (RCL), a novel framework that distills robust supervision from multiple frozen MLLMs into a compact target model. RCL organizes adaptation as a three-stage curriculum that progressively incorporates pseudo-labels based on inter-model agreement and model confidence, enabling stable and noise-aware training. Our approach achieves state-of-the-art performance on standard SFDA datasets, Office-Home, DomainNet-126, and VisDA-C, outperforming zero-shot MLLMs, their ensembles, all without accessing source data or tuning foundation models. Code is available at \href{https://github.com/Dong-Jie-Chen/RCL}{https://github.com/Dong-Jie-Chen/RCL}.
\end{abstract}

\vspace{-4mm}
    
\section{Introduction}

\begin{figure}
    \centering

    \begin{subfigure}[b]{1\linewidth}
        \centering
        \includegraphics[width=0.9\linewidth]{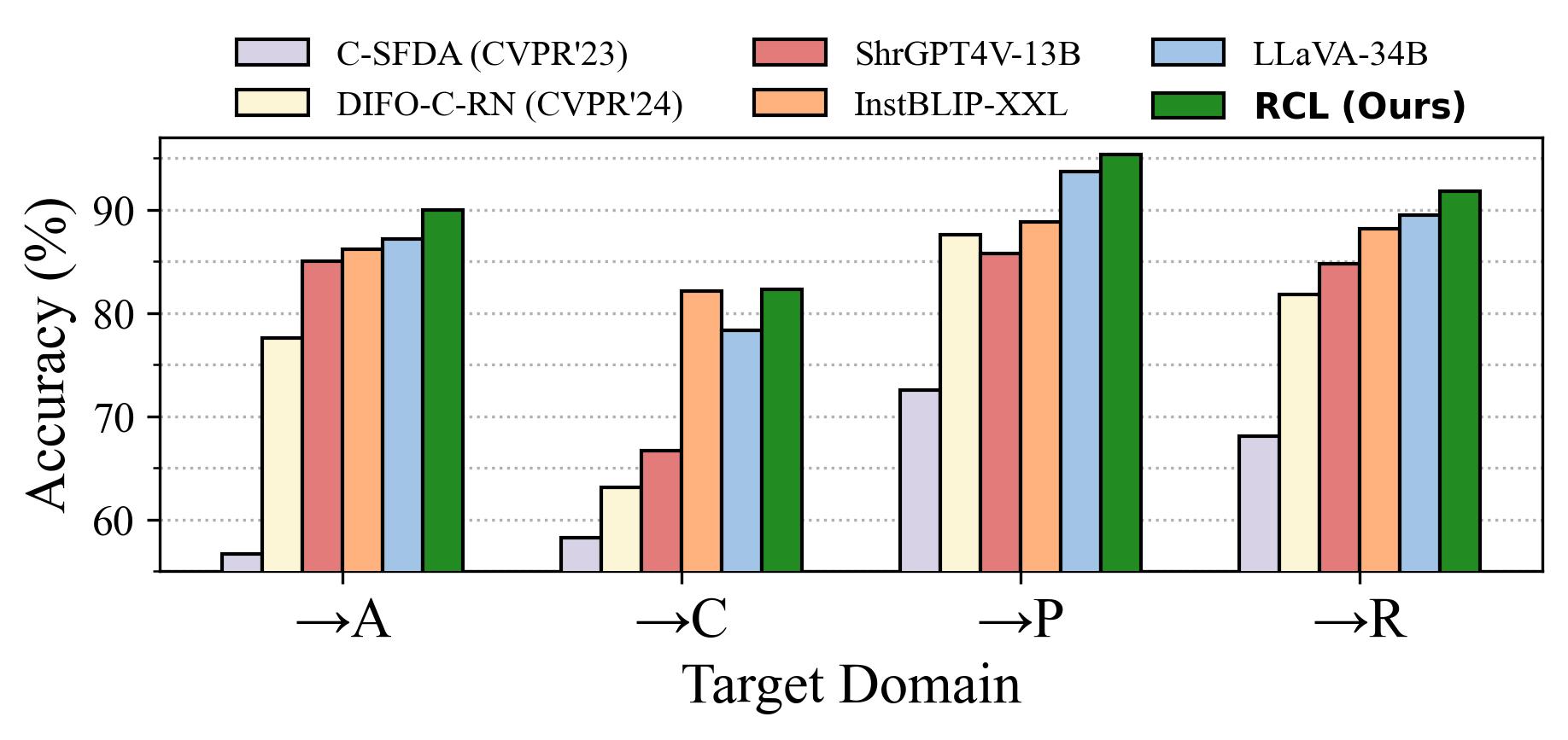}
        \caption{Avg. Accuracy on Office-Home target domains.}
        \label{fig:fig1_a}
    \end{subfigure}
    \vspace{-0.3cm}

    \begin{subfigure}[b]{1\linewidth}
        \centering
        \includegraphics[width=0.9\linewidth]{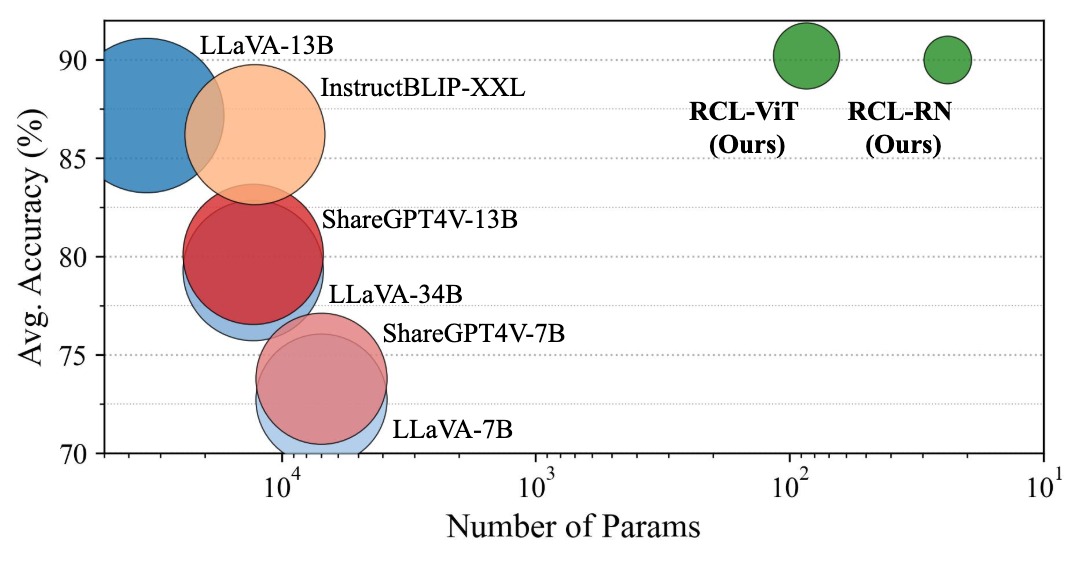}
        \caption{Avg. accuracy vs. number of model parameters (log scale).}
        \label{fig:fig1_b}
    \end{subfigure}
    \vspace{-0.5cm}

    \caption{Comparisons with existing methods, MLLMs (zero-shot with proposed STS), and RCL on OfficeHome dataset. RCL achieves SOTA results across domains while being lightweight.}
    \label{fig:mllm_zero_shot}
    \vspace{-4mm}
\end{figure}

Source-Free Domain Adaptation (SFDA) aims to adapt a pre-trained model to a new target domain without requiring access to labeled source data.
A common trend in SFDA is to use pre-trained models to bridge domain gaps and compensate for the absence of source data~\cite{SHOT_2021, kim2022source, ge2023domain, lai2023padclip, tang2023sourceMLLM}. Earlier works relied on ImageNet-pretrained backbones~\cite{SHOT_2021}, while recent methods incorporate vision-language models (VLMs) like CLIP~\cite{singha2023ad, ge2023domain, lai2023padclip}, using their transferable representations for pseudo-labeling~\cite{tang2023sourceMLLM}, prompt tuning~\cite{lai2023padclip}, or feature alignment~\cite{karim2023c}. However, these approaches typically focus on a single VLM, often involve prompt design or VLM training or finetuning.

While CLIP-based models offer strong visual-language alignment, their contrastive objectives limit semantic depth and reasoning ability. We ask a deeper question: can we move beyond shallow alignment to extract richer, instruction-following supervision from multimodal large language models (MLLMs)—models trained not just to align, but to understand, describe, and reason about images?
To this end, we propose a new paradigm of foundation-model guided SFDA, where we treat MLLMs such as LLaVA~\cite{liu2023improvedllava} as frozen, zero-shot teachers. These models are trained on massive web-scale image-text corpora and encode diverse, high-level knowledge about visual concepts, object interactions, and real-world semantics, often combining high-capacity visual encoders with instruction-tuned LLMs, and recent studies show they now match or surpass CLIP on image classification benchmarks~\cite{liu2024revisiting}. Moreover, domain-specific MLLMs (e.g., medical~\cite{li2023llava-med}) make them a realistic choice when specialized teacher models are absent. 
As shown in Fig.~\ref{fig:fig1_a}, applying MLLMs in a zero-shot manner using our proposed STS (Sec.~\ref{subsec:sts}) module already outperforms SOTA SFDA methods. This validates their ability to generalize across domains without access to the source dataset and provides evidence that MLLMs encode transferable, domain-agnostic visual cues and reasoning that can guide adaptation.

However, directly using MLLMs in deployment is impractical. First, they are generative and not designed for closed-set classification, often violating instructions and producing free-form outputs. Second, as shown in Fig.~\ref{fig:fig1_b}, MLLMs are computationally expensive to query and unsuitable for real-time or on-device applications. Third, different MLLMs produce diverse and occasionally conflicting predictions, making it difficult to trust any single model’s output. This creates a dilemma: while MLLMs contain useful knowledge, their inference behavior is unreliable, and their scale makes them unsuitable for adaptation or downstream use.

To overcome these limitations, we introduce \textbf{Reliability-based Curriculum Learning (RCL)}, a structured, multi-teacher SFDA framework that distills using pseudo-label guidance and self-learning from multiple frozen MLLMs into a compact student model. First, we use our proposed STS module to convert open-ended MLLM outputs into class predictions by computing semantic similarity between generated text and class names. Next, we estimate the reliability of pseudo-labels by measuring agreement across MLLMs, allowing us to partition unlabeled target samples based on consensus.
RCL then trains the student model through a three-stage curriculum. In Reliable Knowledge Transfer (RKT), we use only highly consistent pseudo-labels to initialize adaptation. In Self-correcting and MLLM-guided Knowledge Expansion (SMKE), we incorporate partially ambiguous samples using model confidence and teacher agreement. Finally, Multi-hot Masking Refinement (MMR) addresses noisy supervision by masking low-confidence logits and applying consistency regularization. This staged design ensures stable convergence and maximizes utility from all target samples, while preserving scalability and efficiency. While conventional SFDA approaches rely on pretrained source models, our formulation follows recent trends~\cite{lai2023padclip, tang2023sourceMLLM} that treat foundation models as alternative supervisory signals. By distilling knowledge from multiple frozen MLLMs without requiring source data or finetuning, RCL remains faithful to the SFDA setting while enabling stronger supervision from multimodal sources.

RCL achieves SOTA performance on Office-Home, DomainNet-126, and VisDA-C, outperforming zero-shot MLLM baselines (with STS), majority-vote ensembles, and prior CLIP-based methods. RCL utilizes knowledge from MLLMs (70B to 7B params) into lightweight architectures like ResNet50 (~25M params), achieving up to 280× to 2800× model size reduction while surpassing accuracy (see Fig.~\ref{fig:fig1_b}).
Our key contributions are as follows:

\begin{itemize}
    \item We propose the first SFDA framework that distills knowledge from multiple frozen MLLMs.
    \item We introduce STS, a simple semantic matching method to convert open-ended MLLM outputs into class predictions.
    \item We design RCL, a three-stage curriculum (RKT, SMKE, MMR) that incorporates pseudo-labels based on MLLM agreement and model confidence.
    \item We achieve state-of-the-art results on Office-Home, DomainNet-126, and VisDA-C, outperforming CLIP-based and zero-shot MLLM baselines.
    
\end{itemize}
\section{Related Work}

\textbf{Source-Free Domain Adaptation.} 
SFDA adapts a pre-trained source model to a target domain using only unlabeled target data, making pseudo-labeling central~\cite{qu2022bmd, chenself, litrico2023guiding}. Prior works improve pseudo-labels via target structure~\cite{NRC_2022, tang2021nearest, tang2022semantic, diamant2024confusing, xu2025unraveling}, align distributions~\cite{yang2021generalized, ding2022source}, or generate target-style data~\cite{li2020model, VDMDA_arxiv_2021, 3CGAN_2022}. In contrast, several recent works adapt CLIP in DA/UDA settings with prompt learning, multi-prompt alignment~\cite{chen2023multi}, prompt gradient alignment~\cite{hoang2024enhancing}, prompt learning for DA~\cite{ge2023domain}, and cluster-preserving prompt learning~\cite{vuong2025preserving}. These methods require source data or prompt optimization, whereas we tackle the stricter SFDA setting with single inference on teachers without any modification.

\textbf{VLMs/MLLMs.}  
Pretrained vision-language models (VLMs) like CLIP~\cite{CLIP_2021} and ALIGN~\cite{ALIGN_2021} capture vision-language-aligned features, while recent MLLMs (e.g., LLaVA~\cite{liu2023improvedllava}, InstructBLIP~\cite{dai2024instructblip}) enhance multimodal understanding via LLM backbones, enabling strong zero-shot capabilities. While some works use CLIP for domain adaptation, they require labeled source data~\cite{lai2024empowering, zhao2024learning, westfechtel2023combining, hu2024reclip, zhang2025source} or fine-tuning~\cite{lai2023padclip}.
DIFO~\cite{tang2023sourceMLLM}, the latest SOTA work, adapts CLIP for target-domain learning by iteratively customizing it with prompt learning and distilling its knowledge into a task-specific model. While effective, this approach requires adapting a large VLM, whereas our method leverages multiple MLLMs in a zero-shot manner without finetuning.

\begin{figure*}[!th]
    \centering
    \vspace{-1mm}
    \includegraphics[width=0.85\linewidth]{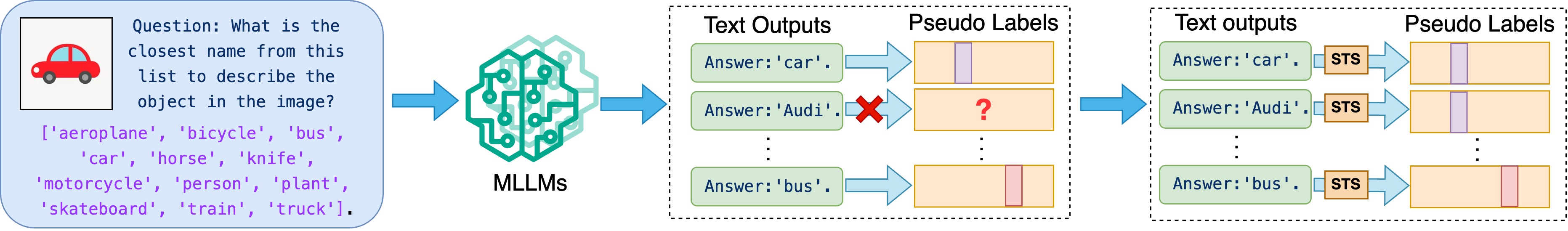}
    \vspace{-0.2cm}
    \caption{Directly prompting MLLMs for classification can lead to failures: we propose semantic textual similarity (STS) in Section~\ref{subsec:sts}.}
    \label{fig:vqa_classification}
\vspace{-0.4cm}
\end{figure*}

 \textbf{Multi-Teacher Knowledge Distillation.}  
MTKD extends KD by combining knowledge from multiple teachers~\cite{jiang2024mtkd, pham2023collaborative}, with strategies including aligning teacher outputs for consistent supervision~\cite{wu2021one}, integrating teacher for better generalization~\cite{li2022mimicking}, using heterogeneous teacher architectures~\cite{kang2023distillation}, attention fusion~\cite{cheng2024lgfa}, or dynamically balancing ensemble size via stochastic teacher selection~\cite{ding2024trade}. While these works highlight the benefits of combining multiple teachers, they generally assume labeled source data, stable teacher ensembles, and access to reliable outputs. RCL addresses the stricter SFDA setting, where no source data is available and teacher predictions (from MLLMs, CLIP, or others) can be inherently noisy. Our framework introduces consensus-based reliability scoring and a staged curriculum to progressively refine pseudo-labels, ensuring robust multi-teacher distillation under uncertainty.

\section{Pseudo-labeling and Reliability}

First, we formally define SFDA for image classification. We denote $\mathcal{D}_s = {(x^{i}_{s},y^{i}_{s})}_{i=1}^{N_s}$ as the labeled source-domain dataset with $N_s$ images, where $x^{i}_{s}\in\mathcal{X}_{s}$ refers to an image and $y^{i}_{s}\in\mathcal{Y}_{s}$ is its corresponding one-hot label. A pre-trained source model $f_{\theta_{s}}:\mathcal{X}_s \rightarrow\mathcal{Y}_s$ is trained on $\mathcal{D}_s$, where $\theta_{s}$ represents its learned parameters.
The target domain contains an unlabeled dataset, $\mathcal{D}_t = \{x_{t}^{i}\}_{i=1}^{N_t}$, where $x_{i}^{t}\in\mathcal{X}_t$ (target domain images), and $N_t$ is the number of unlabeled images.
The goal of SFDA is to adapt a pre-trained source model $f_{\theta_\mathsf{s}}$ to $\mathcal{D}_t$ without access to $\mathcal{D}_s$ during the adaptation process. The goal is to train a target model $f_{\theta_\mathsf{t}}:\mathcal{X}_t \rightarrow\mathcal{Y}_t$, where $\mathcal{Y}_t$ is the target domain label space.

\subsection{Pseudo-labeling with MLLMs using STS}
\label{subsec:sts}

We use multiple MLLMs for initial target-image labeling. As MLLMs are designed primarily for text generation, their responses can deviate from classification. Therefore, we design prompts and reframe VQA as a zero-shot classification task using MLLMs (see Fig.~\ref{fig:vqa_classification}). We design the prompt to incorporate all class names and a question instructing MLLMs to select the most appropriate match.
The prompt and the image $x_{t}^{i}$ are then fed to multiple pre-trained MLLMs, such as LLaVA~\cite{ liu2023improvedllava, liu2023llava}, InstructBLIP~\cite{dai2024instructblip}, and ShareGPT4V~\cite{chen2023sharegpt4v}. 
Each MLLM generates a text output $T_{1}^{i}, T_{2}^{i}, \ldots, T_{M}^{i}$, where $M$ is the number of MLLMs employed.

However, as shown in Fig.~\ref{fig:vqa_classification}, MLLMs sometimes fail to follow prompts, generating responses beyond classification constraints.
This happens when MLLMs rely on prior knowledge rather than selecting from provided options (e.g., predicting ‘Audi’ instead of ‘car’).
Unlike MTKD, which distills structured soft labels, our approach must handle inherently diverse and inconsistent MLLM outputs. To mitigate this, we propose Semantic Textual Similarity (STS) to align outputs with class names, ensuring pseudo-label consistency.
We derive pseudo-labels by computing STS between class names and MLLM-generated text.
Formally, for the $m$-th MLLM and the $i$-th image $x_{t}^{i}$, the pseudo-label $\hat{y}^{mi}$ is determined by
$\hat{y}^{mi} = \operatorname*{argmax}_c \text{STS}(T_{m}^{i}, T_{t}^{c}),$
where $T_{t}^{c}$ represents the name of the $c$-th class. The STS between two text sequences $T_1$ and $T_2$ is computed as:
\begin{equation}
\text{STS}(T_1, T_2) = \frac{\mathbf{v}_1 \cdot \mathbf{v}_2}{\|\mathbf{v}_1\|_2 \|\mathbf{v}_2\|_2},
\end{equation}
where $\mathbf{v}_1$ and $\mathbf{v}_2$ are vector representations of $T_1$ and $T_2$, respectively. 
STS refines pseudo-labels by aligning MLLM outputs with class semantics, even when instructions are ignored. Implementation details and prompt sensitivity analysis are in \textit{Supplementary} (Sec.~\ref{appendix:mllm}).

\begin{figure}[!b]
    \vspace{-4mm}
    \includegraphics[width=0.9\linewidth]{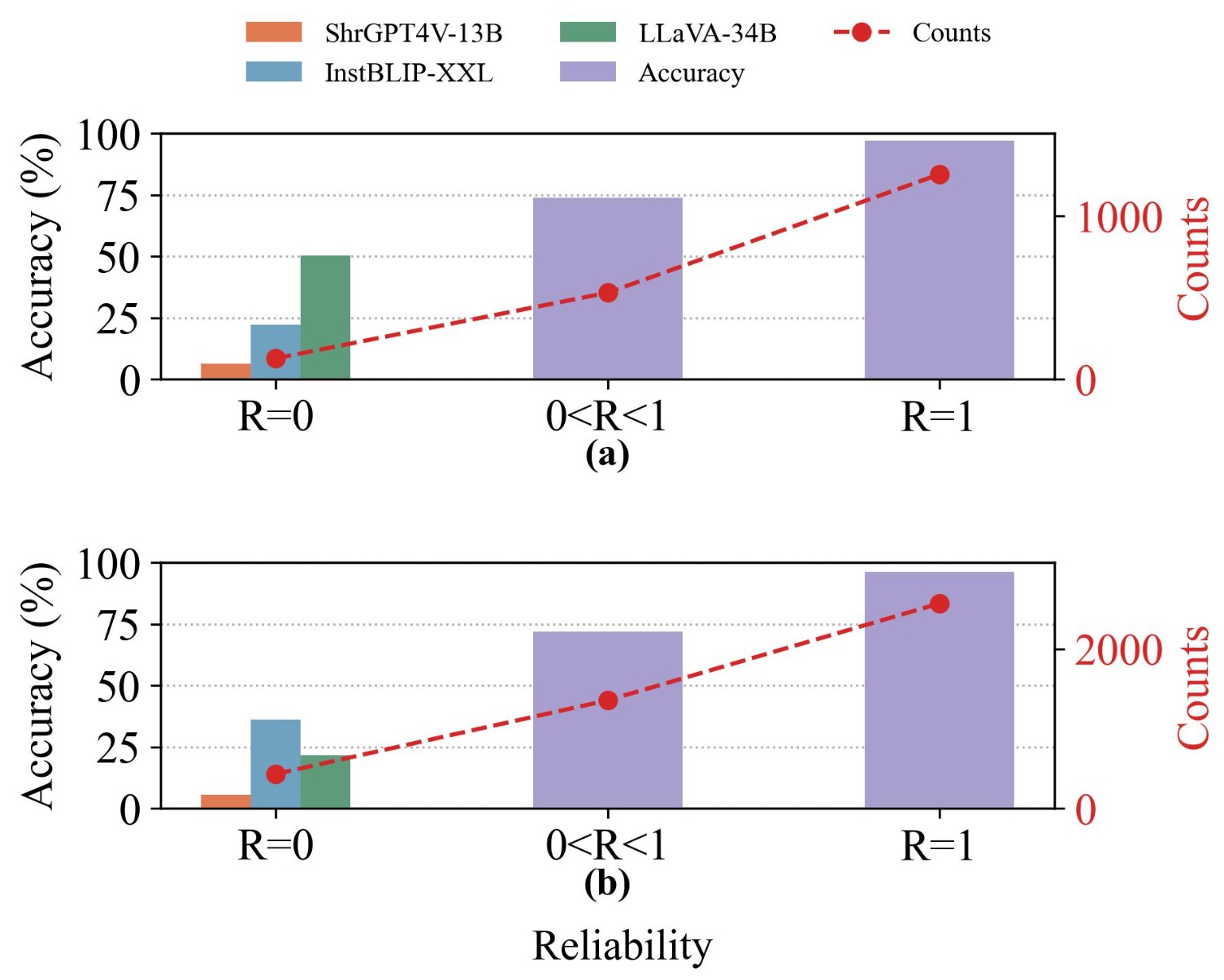}
    \vspace{-4mm}
    \caption{\small{Pseudo-label accuracy and distribution across MLLMs in Office-Home (a) Clipart and (b) Art domains (65 classes each).}}
    \label{fig:reliability}
\vspace{-4mm}
\end{figure}

\begin{figure*}[!htp]
    \centering
    \vspace{-5mm}
    \includegraphics[width=0.85\linewidth]{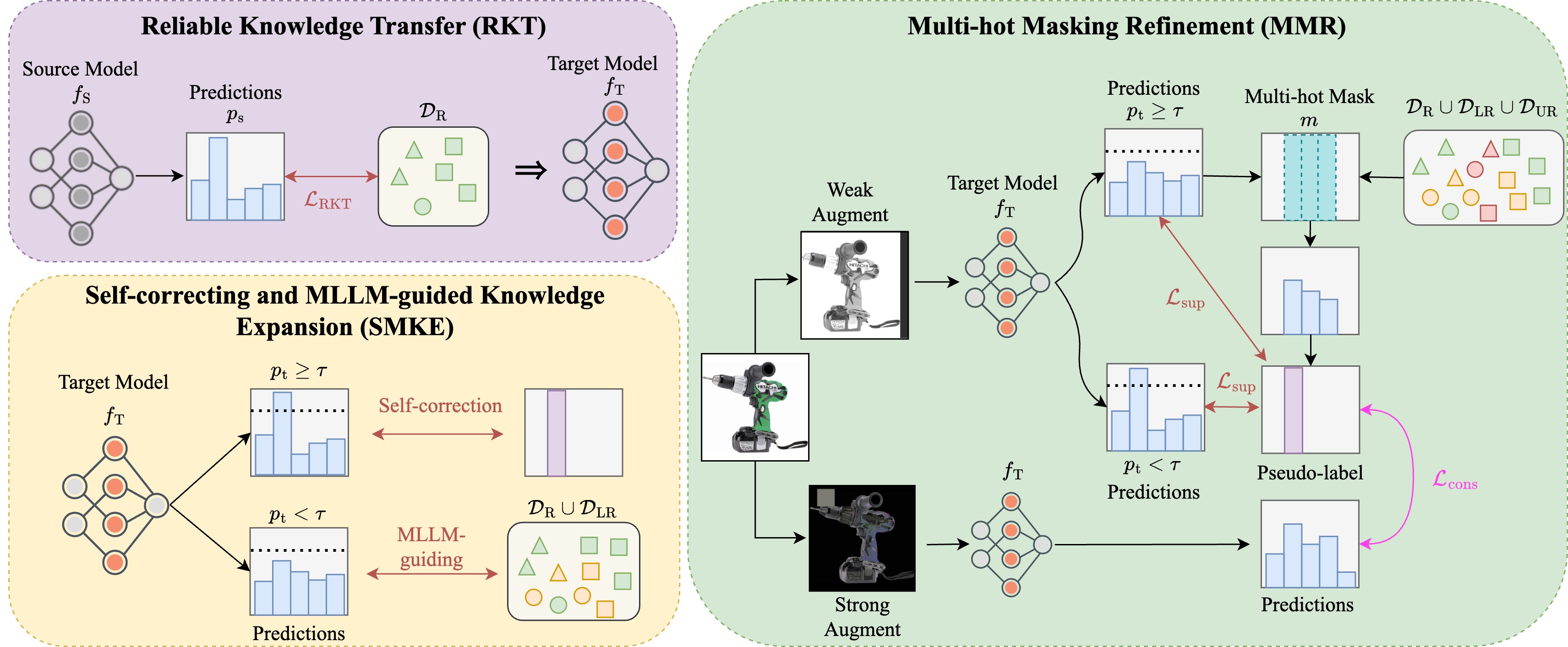}
    \vspace{-0.2cm}
    \caption{Overview of our Reliability-based Curriculum Learning (RCL) framework. RCL applies curriculum learning over target data by progressively incorporating MLLM pseudo-labels based on reliability. (1) \textit{RKT} uses high-confidence samples from full MLLM agreement to initialize feature learning. (2) \textit{SMKE} integrates partially agreed labels to expand knowledge and correct early errors. (3) \textit{MMR} learns from uncertain samples using multi-hot masking and consistency loss. This structured curriculum enables full data use while reducing label noise.}
    \label{fig:overview}
    \vspace{-0.4cm}
\end{figure*} 

\subsection{Consensus-based Reliability Measurement}

Pseudo-labels from different MLLMs may vary for the same target sample.
This disagreement raises a key question:
\textit{how can we measure the reliability of the pseudo-labels from multiple MLLMs?} 
While STS helps correct deviations from instructions, it does not assess pseudo-label reliability, as it cannot detect when MLLMs generate incorrect labels. To address this, we propose a consensus-based reliability metric for pseudo-labels.
We define a reliability score $\mathcal{R}(x_{t}^{i})$ for each target domain sample $x_{t}^{i}$ based on agreement among MLLM-assigned pseudo-labels:
\begin{equation}
\mathcal{R}(x_{t}^{i}) = \frac{1}{M(M-1)} \sum_{m=1}^{M} \sum_{n=1, n \neq m}^{M} \mathbbm{1}(\hat{y}^{mi} = \hat{y}^{ni}),
\end{equation}

\noindent where $\mathbbm{1}(\cdot)$ is the indicator function. $\mathcal{R}(x_{t}^{i})$ quantifies the proportion of MLLM pairs that assign the same pseudo-label to $x_{t}^{i}$. 
Using $\mathcal{R}(x_{t}^{i})$, we categorize $\mathcal{D}_t$ into three subsets:  
(1) \textit{Reliable} (\(\mathcal{D}_\textit{R}\)): All MLLMs agree (\(\mathcal{R}(x_{t}^{i}) = 1\)),  
(2) \textit{Less Reliable} (\(\mathcal{D}_\textit{LR}\)): Partial agreement (\(0 < \mathcal{R}(x_{t}^{i}) < 1\)), and  
(3) \textit{Unreliable} (\(\mathcal{D}_\textit{UR}\)): No agreement (\(\mathcal{R}(x_{t}^{i}) = 0\)).
Thus, the target dataset is partitioned as:  $\mathcal{D}_t = \{\mathcal{D}_\textit{R}, \mathcal{D}_\textit{LR}, \mathcal{D}_\textit{UR}\}.$

We observe that higher reliability scores correlate with improved pseudo-label accuracy, validating our consensus-based reliability metric. Fig.~\ref{fig:reliability} further shows pseudo-label accuracy and sample distribution across reliability levels $\mathcal{R}(x_{t}^{i})$, highlighting disagreement among MLLMs. For samples with $\mathcal{R}(x_{t}^{i}) > 0$, accuracy is computed via majority vote across MLLMs, while for $\mathcal{R}(x_{t}^{i}) = 0$, we report the accuracy of each individual MLLM.
\vspace{-2mm}

\section{Reliability-based Curriculum Learning}

Pseudo-labels from teachers provide a strong starting point, but their reliability can vary across samples. 
To address this, we propose Reliability-based Curriculum Learning (RCL) (Fig.~\ref{fig:overview}), which uses consensus-based reliability to progressively incorporate target data by pseudo-label confidence.  
RCL consists of three stages: (1) \textit{Reliable Knowledge Transfer (RKT)} trains on the most reliable subset $\mathcal{D}_\textit{R}$; (2) \textit{Self-correcting and MLLM-guided Knowledge Expansion (SMKE)} integrates less reliable samples with adaptive correction; and (3) \textit{Multi-hot Masking Refinement (MMR)} incorporates the full target set $\mathcal{D}_t$, using a union-of-classes mask with consistency regularization to handle the most uncertain subset $\mathcal{D}_\textit{UR}$. This staged curriculum allows the student to learn from confident samples first and gradually expand to noisier ones, stabilizing adaptation.  
RCL is teacher-agnostic, being a multi-teacher curriculum designed to handle noisy and unreliable pseudo-labels. This makes the framework broadly applicable to CLIP or other models. Our STS conversion integrates MLLMs by adapting their free-form outputs into usable class predictions. We view MLLMs as an upper-bound case: their foundation knowledge and demonstrated cross-domain generalization offer a strong motivation for SFDA in real-world settings such as medical applications.

\subsection{Reliable Knowledge Transfer (RKT)}
\label{sec:rkt}

In the first stage, we train the student on the reliable subset $\mathcal{D}_\textit{R}$ where multiple teachers fully agree.  
This ensures that early learning is guided by high-quality labels and avoids propagating noise.  
\textit{The intuition is to give the student a “clean start”: by only learning from samples with consensus, the model builds a stable foundation before being exposed to more uncertain data.}  
The reliable subset $\mathcal{D}_\textit{R}$ consists of samples for which all MLLMs agree on the pseudo-label 
$\mathcal{D}_\textit{R} = \{(x^i_r, y^i_r) \mid \mathcal{R}(x^i_r) = 1\},$
where $x^i_r$ is the $i$-th sample in the reliable subset, $\tilde{y}^i_r$ is the corresponding pseudo-label agreed upon by all MLLMs, and $\mathcal{R}(x^i_r) \in [0,1]$ is the reliability measure defined in the previous section.
The target model $f_{\theta_t}$ is trained using a supervised cross-entropy loss on the reliable subset $\mathcal{D}_\textit{R}$:
\vspace{-1mm}
\begin{equation}
\mathcal{L}_{\textit{RKT}} = -\frac{1}{|\mathcal{D}_R|} \sum_{(x^i_r, y^i_r) \in \mathcal{D}_\textit{R}} y^i_r \cdot \log f_{\theta_t}(x^i_r),
\end{equation}

\noindent where $|\mathcal{D}_R|$ denotes the number of samples in the reliable subset.
Training exclusively on $\mathcal{D}_R$ ensures that only the most confident and consistent MLLM pseudo-labels shape the target model’s initial learning phase. RKT provides a strong foundation before introducing less reliable pseudo-labels, ensuring stable knowledge transfer.

\subsection{Self-correcting and MLLM-guided Knowledge Expansion (SMKE)} 
\label{sec:smke}

Following RKT, RCL expands training by incorporating less reliable pseudo-labels to broaden the student’s knowledge. Since the student has already been pre-trained and stabilized through RKT, we transition from strict teacher supervision to a more adaptive learning process. When the student is confident, it self-corrects its predictions; when confidence is low, MLLMs provide auxiliary guidance rather than acting as fixed teachers.
\textit{The intuition is that after RKT, the student has matured enough to act as a co-teacher: its predictions can be trusted to correct or supplement partial teacher agreement, enabling a gradual and safer expansion of usable data.}  
To implement this, SMKE fine-tunes the model on both reliable and less reliable subsets, $\mathcal{D}_R \cup \mathcal{D}_{LR}$. For each sample $x_t^i$, the pseudo-label $\tilde{y}^i$ is determined adaptively: if the student’s prediction $\hat{y}_t^i$ has confidence $p_t^i$ above a threshold $\tau$, we use it; otherwise, we defer to the MLLM teachers’ pseudo-label. This balances the student’s growing reliability with external guidance, reducing dependence on any single source.
The pseudo-label $\tilde{y}^i$ is  as follows:
\begin{equation}
\tilde{y}^i =
\begin{cases}
\hat{y}_t^i, & \text{if } p_t^i \geq \tau, \\
\operatorname*{mode}({\hat{y}^{1i}, \hat{y}^{2i}, \ldots, \hat{y}^{Mi}}), & \text{if } p_t^i < \tau ,
\end{cases}
\end{equation}
where $\operatorname*{mode}(\cdot)$ returns the most frequent pseudo-label among the MLLMs.

In SMKE, if the target model's confidence score $p_t^i > \tau$, ($\tau$ being given threshold), we employ the target model's pseudo-label $\hat{y}_t^i$ for self-correction. 
Otherwise, the model adopts the most frequent MLLM pseudo-label to mitigate uncertainty and expand knowledge.
The adaptive training approach is optimized through the loss function:
\begin{equation}
    \mathcal{L}_\text{SMKE} = -\frac{1}{|\mathcal{D}_\textit{R} \cup \mathcal{D}_\textit{LR}|}  \sum_{x^i_t \in \{ \mathcal{D}_\textit{R} \cup \mathcal{D}_\textit{LR}\}} \tilde{y}^i \cdot \log f_{\theta_t}(x^i_t).
\end{equation}

By incorporating the less reliable pseudo-labels and leveraging the target model's confidence scores, SMKE stage of the curriculum learning framework can expand the knowledge transferred to the target model by utilizing both the target model's predictions and the MLLMs' pseudo-labels. 

\subsection{Multi-hot Masking Refinement (MMR)} 
\label{sec:mmr}

Finally, we refine the student on the full target dataset $\mathcal{D}_t$, focusing on the most uncertain subset $\mathcal{D}_\textit{UR}$.  
Naively minimizing entropy on these samples risks overfitting to noise or distractors.  
\textit{MMR instead narrows the prediction space by masking logits to the union of teacher-suggested classes, forcing the model to choose only among classes deemed plausible by at least one teacher.}  
We further apply consistency regularization between weak and strong augmentations to stabilize training when agreement is low.  
\textit{The key intuition is that by this stage the student has already learned reliable patterns (via RKT and SMKE), so its own predictions are valuable. MMR leverages both the student’s maturity and the teachers’ broad guidance: disagreement is treated as ambiguity rather than error, and the mask provides a constrained search space for refinement.}  
This enables the model to still benefit from hard samples while reducing error propagation.

\begin{table*}[t!]
\vspace{-0.3cm}
\caption{\label{tab:office-home} Accuracy (\%) on \textbf{Office-Home}. SF: source-free; CP, ViT: CLIP, ViT backbone. Best results are bolded, second-best underlined. (*) indicates zero-shot performance. Full table in \textit{Supplementary Material}, Tab.~\ref{tab:officehome-full}.} 
\renewcommand\tabcolsep{2.3pt}  
\renewcommand\arraystretch{1.0}  
\centering
\scriptsize
\resizebox{\textwidth}{!}{  
    \begin{tabular}{l|ccc|cccccccccccc|c}
    \toprule
    Method & SF & CP & ViT & A$\rightarrow$C & A$\rightarrow$P & A$\rightarrow$R & C$\rightarrow$A & C$\rightarrow$P & C$\rightarrow$R & P$\rightarrow$A & P$\rightarrow$C & P$\rightarrow$R & R$\rightarrow$A & R$\rightarrow$C & R$\rightarrow$P & Avg.\\
    \midrule
    Source & - & \textcolor{red}{\xmark} & \textcolor{red}{\xmark} & 44.7 & 64.2 & 69.4 & 48.3 & 57.9 & 60.3 & 49.5 & 40.3 & 67.2 & 59.7 & 45.6 & 73.0 & 56.7 \\
    \midrule
    PADCLIP-RN~\cite{lai2023padclip} & \textcolor{red}{\xmark} & \textcolor{Green}{\cmark} & \textcolor{red}{\xmark} & 57.5 & 84.0 & 83.8 & 77.8 & 85.5 & 84.7 & 76.3 & 59.2 & 85.4 & 78.1 & 60.2 & 86.7 & 76.6 \\
    ADCLIP-RN~\cite{singha2023ad} & \textcolor{red}{\xmark} & \textcolor{Green}{\cmark} & \textcolor{red}{\xmark} & 55.4 & 85.2 & 85.6 & 76.1 & 85.8 & 86.2 & 76.7 & 56.1 & 85.4 & 76.8 & 56.1 & 85.5 & 75.9 \\
    \midrule
    ELR~\cite{yi2023source} & \textcolor{Green}{\cmark} & \textcolor{red}{\xmark} & \textcolor{red}{\xmark} & 58.4 & 78.7 & 81.5 & 69.2 & 79.5 & 79.3 & 66.3 & 58.0 & 82.6 & 73.4 & 59.8 & 85.1 & 72.6 \\
    PLUE~\cite{litrico_2023_CVPR} & \textcolor{Green}{\cmark} & \textcolor{red}{\xmark} & \textcolor{red}{\xmark} & 49.1 & 73.5 & 78.2 & 62.9 & 73.5 & 74.5 & 62.2 & 48.3 & 78.6 & 68.6 & 51.8 & 81.5 & 66.9 \\
    C-SFDA~\cite{karim2023c} & \textcolor{Green}{\cmark} & \textcolor{red}{\xmark} & \textcolor{red}{\xmark} & 60.3 & 80.2 & 82.9 & 69.3 & 80.1 & 78.8 & 67.3 & 58.1 & 83.4 & 73.6 & 61.3 & 86.3 & 73.5 \\
    PSAT-GDA~\cite{tang2023progressive} & \textcolor{Green}{\cmark} & \textcolor{red}{\xmark} & \textcolor{Green}{\cmark} & 73.1 & 88.1 & 89.2 & 82.1 & 88.8 & 88.9 & 83.0 & 72.0 & 89.6 & 83.3 & 73.7 & 91.3 & 83.6 \\
    \midrule
    DIFO-C-RN~\cite{tang2023sourceMLLM} & \textcolor{Green}{\cmark} & \textcolor{Green}{\cmark} & \textcolor{red}{\xmark} & 62.6 & 87.5 & 87.1 & 79.5 & 87.9 & 87.4 & 78.3 & 63.4 & 88.1 & 80.0 & 63.3 & 87.7 & 79.4 \\
    DIFO-C-B32~\cite{tang2023sourceMLLM} & \textcolor{Green}{\cmark} & \textcolor{Green}{\cmark} & \textcolor{Green}{\cmark} & 70.6 & 90.6 & 88.8 & 82.5 & 90.6 & 88.8 & 80.9 & 70.1 & 88.9 & 83.4 & 70.5 & 91.2 & 83.1 \\
    \midrule
    CLIP-RN~\cite{CLIP_2021}* & - & \textcolor{Green}{\cmark} & \textcolor{red}{\xmark} & 51.7 & 85.0 & 83.7 & 69.3 & 85.0 & 83.7 & 69.3 & 51.7 & 83.7 & 69.3 & 51.7 & 85.0 & 72.4 \\
    LLaVA-34B (w/ STS)~\cite{liu2023llava}* & - & \textcolor{Green}{\cmark} & \textcolor{Green}{\cmark} & 78.3 & 93.7 & 89.5 & 87.0 & 93.7 & 89.5 & 87.0 & 78.3 & 89.5 & 87.0 & 78.3 & 93.7 & 87.2 \\
    InstBLIP-XXL (w/ STS)~\cite{dai2024instructblip}* & - & \textcolor{Green}{\cmark} & \textcolor{Green}{\cmark} & 82.0 & 91.6 & 88.8 & 82.2 & 91.6 & 88.8 & 82.2 & 82.0 & 88.8 & 82.2 & 82.0 & 91.6 & 86.2 \\
    ShrGPT4V-13B (w/ STS)~\cite{chen2023sharegpt4v}* & - & \textcolor{Green}{\cmark} & \textcolor{Green}{\cmark} & 66.7 & 85.8 & 84.8 & 83.2 & 85.8 & 84.8 & 83.2 & 66.7 & 84.8 & 83.2 & 66.7 & 85.8 & 80.1 \\
    \midrule
    \textbf{RCL (Ours)} & \textcolor{Green}{\cmark} & \textcolor{red}{\xmark} & \textcolor{red}{\xmark} & \underline{82.5} & \underline{95.3} & \textbf{93.3} & \underline{89.1} & \textbf{95.3} & \textbf{92.7} & \textbf{89.3} & \textbf{82.4} & \underline{92.8} & \underline{89.4} & \underline{82.1} & \underline{95.4} & \underline{90.0} \\
    \textbf{RCL-ViT (Ours)} & \textcolor{Green}{\cmark} & \textcolor{red}{\xmark} & \textcolor{Green}{\cmark} & \textbf{83.1} & \textbf{95.7} & \underline{93.1} & \textbf{89.2} & \textbf{95.3} & \underline{92.6} & \underline{89.2} & \underline{82.3} & \textbf{92.9} & \textbf{90.0} & \textbf{83.2} & \textbf{95.5} & \textbf{90.2} \\
    \bottomrule
    \end{tabular}}
    \vspace{-0.3cm}
\end{table*}

\begin{table*}[t]
\caption{\label{tab:domainnet} Accuracy (\%) on \textbf{DomainNet} and average accuracy on \textbf{VisDA}. Full table in \textit{Supplementary Material}, Tab.~\ref{tab:domainnet-full}, ~\ref{tab:visda-full}.} 
\vspace{-2mm}
\renewcommand\tabcolsep{2.3pt}  
\renewcommand\arraystretch{1.0}  
\centering
\scriptsize
\resizebox{\textwidth}{!}{  
    \begin{tabular}{l|ccc|cccccccccccc|c|c}
    \toprule
    \multirow{2}{*}{Method} & \multirow{2}{*}{SF} & \multirow{2}{*}{CP} & \multirow{2}{*}{ViT}  &\multicolumn{13}{c}{\textbf{DomainNet}}\vline
        &{\textbf{VisDA}} \\
        & & & & C$\rightarrow$P & C$\rightarrow$R & C$\rightarrow$S & P$\rightarrow$C & P$\rightarrow$R & P$\rightarrow$S & R$\rightarrow$C & R$\rightarrow$P & R$\rightarrow$S & S$\rightarrow$C & S$\rightarrow$P & S$\rightarrow$R & Avg. & S$\rightarrow$R\\
    \midrule
    Source & - & \textcolor{red}{\xmark} & \textcolor{red}{\xmark} & 42.6 & 53.7 & 51.9 & 52.9 & 66.7 & 51.6 & 49.1 & 56.8 & 43.9 & 60.9 & 48.6 & 53.2 & 52.7 & 45.3 \\
    \midrule
    DAPL-RN~\cite{ge2023domain} & \textcolor{red}{\xmark} & \textcolor{Green}{\cmark} & \textcolor{red}{\xmark} & 72.4 & 87.6 & 65.9 & 72.7 & 87.6 & 65.6 & 73.2 & 72.4 & 66.2 & 73.8 & 72.9 & 87.8 & 74.8 & 86.9 \\
    ADCLIP-RN~\cite{lai2023padclip} & \textcolor{red}{\xmark} & \textcolor{Green}{\cmark} & \textcolor{red}{\xmark} & 71.7 & 88.1 & 66.0 & 73.2 & 86.9 & 65.2 & 73.6 & 73.0 & 68.4 & 72.3 & 74.2 & 89.3 & 75.2 & 88.5  \\
    \midrule
    PLUE~\cite{litrico_2023_CVPR} & \textcolor{Green}{\cmark} & \textcolor{red}{\xmark} & \textcolor{red}{\xmark} & 59.8 & 74.0 & 56.0 & 61.6 & 78.5 & 57.9 & 61.6 & 65.9 & 53.8 & 67.5 & 64.3 & 76.0 & 64.7 & 88.3 \\
    TPDS~\cite{tang2023source} & \textcolor{Green}{\cmark} & \textcolor{red}{\xmark} & \textcolor{red}{\xmark} & 62.9 & 77.1 & 59.8 & 65.6 & 79.0 & 61.5 & 66.4 & 67.0 & 58.2 & 68.6 & 64.3 & 75.3 & 67.1 & 87.6 \\
    \midrule
    DIFO-C-RN~\cite{tang2023sourceMLLM} & \textcolor{Green}{\cmark} & \textcolor{Green}{\cmark} & \textcolor{red}{\xmark} & 73.8 & 89.0 & 69.4 & 74.0 & 88.7 & 70.1 & 74.8 & 74.6 & 69.6 & 74.7 & 74.3 & 88.0 & 76.7 & 88.8 \\
    DIFO-C-B32~\cite{tang2023sourceMLLM} & \textcolor{Green}{\cmark} & \textcolor{Green}{\cmark} & \textcolor{Green}{\cmark} & 76.6 & 87.2 & 74.9 & 80.0 & 87.4 & 75.6 & 80.8 & 77.3 & 75.5 & 80.5 & 76.7 & 87.3 & 80.0 & 90.3\\
    \midrule
    LLaVA-34B (w/ STS)~\cite{liu2023llava}* & - & \textcolor{Green}{\cmark} & \textcolor{Green}{\cmark} & 84.4 & 91.0 & 83.7 & 85.5 & 91.0 & 83.7 & 85.5 & 84.4 & 83.7 & 85.5 & 84.4 & 91.0 & 86.1 & 92.1 \\
    InstBLIP-XXL (w/ STS)~\cite{dai2024instructblip}* & - & \textcolor{Green}{\cmark} & \textcolor{Green}{\cmark} & 82.5 & 89.0 & 83.0 & 86.7 & 89.0 & 83.0 & 86.7 & 82.5 & 83.0 & 86.7 & 82.5 & 89.0 & 85.3 & 86.7 \\
    ShrGPT4V-13B (w/ STS)~\cite{chen2023sharegpt4v}* & - & \textcolor{Green}{\cmark} & \textcolor{Green}{\cmark} & 79.7 & 87.9 & 79.2 & 79.9 & 87.9 & 79.2 & 79.9 & 79.7 & 79.2 & 79.9 & 79.7 & 87.9 & 81.7 & 90.4 \\
    \midrule
    \textbf{RCL (Ours)} & \textcolor{Green}{\cmark} & \textcolor{red}{\xmark} & \textcolor{red}{\xmark} & \underline{87.6} & \underline{92.8} & \underline{87.9} & \underline{89.2} & \underline{92.7} & \underline{87.8} & \underline{89.6} & \underline{87.7} & \underline{87.6} & \underline{89.4} & \underline{87.5} & \underline{92.7} & \underline{89.4} & \underline{93.2} \\
    \textbf{RCL-ViT (Ours)} & \textcolor{Green}{\cmark} & \textcolor{red}{\xmark} & \textcolor{Green}{\cmark} & \textbf{88.1} & \textbf{93.3} & \textbf{88.0} & \textbf{89.7} & \textbf{93.3} & \textbf{88.0} & \textbf{89.7} & \textbf{88.0} & \textbf{87.8} & \textbf{89.7} & \textbf{88.1} & \textbf{93.3} & \textbf{89.7} & \textbf{93.3} \\
    \bottomrule
    \end{tabular}}
\vspace{-4mm}
\end{table*}

\textbf{Multi-hot masking.} Let $\mathbf{z}_t^i \in \mathbb{R}^C$ be the predictive probabilities of the target model for the target domain sample $x_t^i$, where $C$ is the number of classes. The confidence score of the target model's prediction is given by $p_t^i=\operatorname*{max}_c {\mathbf{z}_t^i}$. We define a Multi-hot mask $\mathbf{m}^i \in \{0, 1\}^C$ based on the pseudo-labels assigned by the MLLMs: $\mathbf{m}^i = 1 - \prod_{m=1}^M (1 - \mathbbm{1}(\hat{y}^{mi}))$
where the mask $\mathbf{m}^i$ is formed by adding up one-hot vectors indicating the presence of each class as predicted by the MLLMs for the sample $x_t^i$. We then apply the Multi-hot mask to mask out the target model's logits, forming a refined pseudo-label $\tilde{y}^i$ based on the confidence threshold $\tau$:
\begin{equation}
\tilde{y}^i =
\begin{cases}
\operatorname*{argmax}_C(\mathbf{z}_t^i), & \text{if } p_t^i \geq \tau, \\
\operatorname*{argmax}_C(\mathbf{z}_t^i \odot \mathbf{m}^i), & \text{if } p_t^i < \tau,
\end{cases}
\end{equation}
where $\odot$ denotes element-wise multiplication and $\tau$ is the confidence threshold. If $p_t^i > \tau$, the original prediction is retained; otherwise, it is adjusted based on the multi-hot mask, filtering out less likely classes.
\begin{equation}
    \mathcal{L}_\text{sup} = -\frac{1}{|\mathcal{D}_\textit{R} \cup \mathcal{D}_\textit{LR} \cup \mathcal{D}_\textit{UR}|}  \sum_{x^i_t \in \{\mathcal{D}_\textit{R} \cup \mathcal{D}_\textit{LR} \cup \mathcal{D}_\textit{UR}\}} \tilde{y}^i \cdot \log f_{\theta_t}(x^i_t),
\end{equation}
The consistency loss $\mathcal{L}_{\text{cons}}$ is computed using the refined pseudo-labels from both weakly and strongly augmented samples, reinforcing target model predictions to align with MLLMs, especially when the model is not confident:
\begin{equation}
\mathcal{L}_{\text{cons}} = \frac{1}{M} \sum_{m=1}^{M} \sum_{i=1}^{N_t}  \mathcal{L}_{\text{CE}}(\tilde{y}^i, \mathbf{z}_{st}^i),
\end{equation}
where $\mathbf{z}_{st}^i$ denotes the target model's logit for strong augmentation samples and $\mathcal{L}_{\text{CE}}(\cdot, \cdot)$ denotes the cross-entropy loss. The target model is then optimized through the combined loss $\mathcal{L}_{\text{MMR}} = \mathcal{L}_{\text{sup}} + \lambda_\text{cons}\mathcal{L}_{\text{cons}}$
where $\lambda_\text{cons}$ is a fixed hyperparameter to balance the supervised and consistency losses. A detailed algorithm for MMR is present in \textit{Supplementary}.
Through the MMR phase, the target model not only uses the MLLMs' pseudo-labels to refine its training strategy but also ensures robust learning even from samples whose initial predictions lack confidence.

\section{Experiments}

\textbf{Datasets.}
We evaluate our method on three standard benchmark datasets: Office-Home, DomainNet-126, and VisDA-C 2017. 
\textit{Office-Home}~\cite{saenko2010adapting} has 4 domains -- Real (R), Clipart (C), Art (A), and Product (P), encompassing 65 classes with a total of 15.5k images.
\textit{VisDA}~\cite{peng2017visda} is a large-scale synthetic-to-real object recognition dataset, where the source domain includes 152k synthetic images and the target domain contains about 55k real object images across 12 classes.
\textit{DomainNet}~\cite{peng2019moment} is a challenging large-scale, with 6 domains (around 600k images across 345 classes). We follow the standard setup with 145k images from 126 classes, sampled from four domains, Clipart (C), Painting (P), Real (R), Sketch (S).

\noindent \textbf{Model details.}
Following~\cite{tang2023sourceMLLM, SHOT_2021, tang2023progressive, tang2023source}, we use ResNet-101~\cite{he2016deep} for VisDA and ResNet-50 for Office-Home, DomainNet. For VisDA and Office-Home, we adopt pre-trained source models from SHOT~\cite{SHOT_2021}, while for DomainNet, we train source models following~\cite{litrico_2023_CVPR}. 
For main results, we use the three strongest open-source MLLMs: LLaVA-v1.6-34B~\cite{liu2023llava}, ShareGPT4V-13B~\cite{chen2023sharegpt4v}, and InstructBLIP-T5-XXL~\cite{dai2024instructblip}. RCL training is lightweight and only requires forward passes through frozen MLLMs (used once to cache pseudo-labels). We use NVIDIA A100 GPUs for all adaptation and MLLM inference due to availability. Additional training and model details are in \textit{Supplementary}.

\begin{figure*}[!t]
\vspace{-0.3cm}
    \centering
    \includegraphics[width=0.85\linewidth]{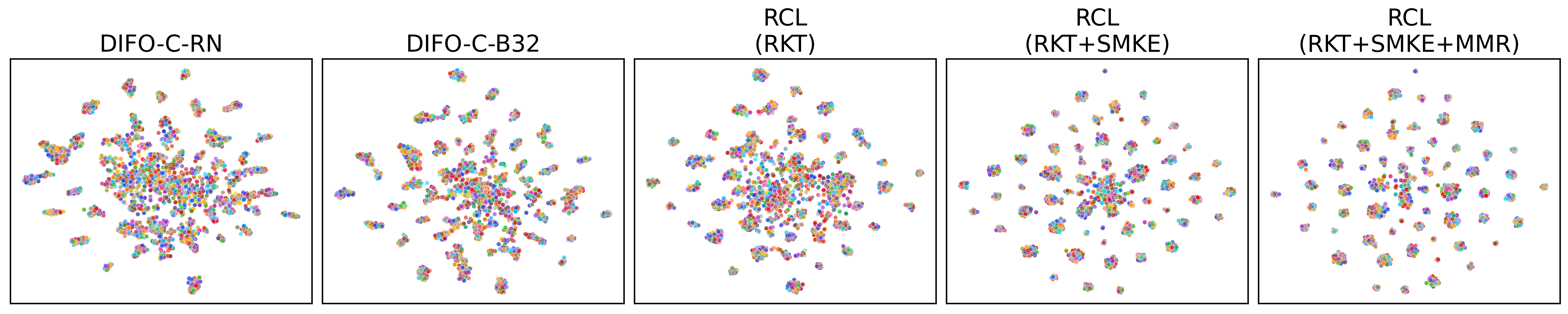}
    \vspace{-0.3cm}
   \caption{t-SNE feature distribution for A$\rightarrow$C in Office-Home. DIFO-C-B32 uses ViT-B32; others use ResNet-50.}
    \label{fig:tsne}
\vspace{-0.4cm}
\end{figure*}

\subsection{Main Results}\label{subsec:main_results}

Tables~\ref{tab:office-home} and~\ref{tab:domainnet} present our results for Office-Home, DomainNet, and VisDA. From top to bottom, we show domain adaptation methods using source data with CLIP-based techniques, followed by source-free methods without multimodal or CLIP, and finally, CLIP-based multimodal methods. Zero-shot MLLM (w/ STS)/CLIP performance is also provided as a reference. \emph{RCL consistently achieves best performance across all datasets, showing notable improvements on existing SOTA: +6.4\% on Office-Home, +9.4\% on DomainNet, and +2.9\% on VisDA-C}. 
The prior best-performing methods (DIFO-C-B32, DIFO-C-RN~\cite{tang2023sourceMLLM}) use a CLIP encoder with ViT-B/32 and a ResNet CLIP backbone, respectively, and use prompt learning. In contrast, \emph{RCL outperforms these using only the ResNet backbone with guided curriculum training through MLLM pseudo-labels, without prompt learning}. Our self-refinement and curriculum learning processes surpass MLLM performance, highlighting the curriculum learning process’s ability to capture valuable latent information in the target domain beyond MLLMs. 
Unlike prior works, our method utilizes zero-shot MLLM inference capability with STS and \emph{does not require customization, prompt learning, or heavy training of multimodal models}.

\subsection{Ablation Study}

We conduct ablation studies to analyze the contributions of RCL’s core components, the role of teacher pseudo-labels (with and without MLLMs), robustness to MLLM strength and ensemble size, and knowledge transfer to smaller student architectures. These insights help disentangle where RCL’s improvements originate. Additional results, ablations, and analysis are provided in the \textit{Supplementary}.

\begin{table}[!h]
\caption{Ablation study on the impact of RCL components.}
\vspace{-2mm}
\centering
\scalebox{0.9}{ 
\begin{tabular}{ccc|ccccc}
\toprule
\multicolumn{3}{c|}{RCL} & \multicolumn{5}{c}{Office-Home} \\
RKT & SMKE & MMR & $\rightarrow$A & $\rightarrow$C & $\rightarrow$P & $\rightarrow$R & \multicolumn{1}{|c}{Avg.} \\ \midrule
\textcolor{Green}{\cmark} & \textcolor{red}{\xmark} & \textcolor{red}{\xmark} & 82.8 & 73.3 & 89.3 & 88.1 & 83.3 \\
\textcolor{Green}{\cmark} & \textcolor{red}{\xmark} & \textcolor{Green}{\cmark} & 87.7 & 80.2 & 93.3 & 92.0 & 88.3 \\
\textcolor{Green}{\cmark} & \textcolor{Green}{\cmark} & \textcolor{red}{\xmark} & 88.5 & 80.9 & 95.1 & 92.5 & 89.3 \\
\textcolor{Green}{\cmark} & \textcolor{Green}{\cmark} & \textcolor{Green}{\cmark} & \textbf{89.3} & \textbf{82.3} & \textbf{95.3} & \textbf{92.9} & \textbf{90.0} \\
\bottomrule
\end{tabular}%
}
\label{tab:rcl_comp_ablation}
\end{table}

\textbf{Impact of RCL components.} 
Table~\ref{tab:rcl_comp_ablation} shows the results of evaluating individual RCL components. Using only RKT yields the lowest performance, as it relies on the most reliable MLLM pseudo-labels, which may lack class coverage and diversity (see \textit{Supplementary} Fig.~\ref{fig:class_dist_officehome}). \emph{RKT still provides essential initial supervision for feature grounding.} Adding SMKE outperforms applying MMR directly after RKT, as \emph{SMKE expands knowledge by incorporating less reliable pseudo-labels, improving robustness}. In contrast, MMR without intermediate refinement underperforms, highlighting the need for pseudo-label diversity before semi-supervised learning. Finally, \emph{MMR after SMKE consistently improves performance by enabling learning from even the most unreliable labels, maximizing dataset utilization.} Fig.~\ref{fig:tsne} visualizes how RCL progressively refines target features compared to SOTA methods like DIFO.

\begin{table}[ht]
\caption{Performance of RCL without MLLMs on the Office-Home dataset. RCL uses three teacher models $A, B, C$.}
\vspace{-2mm}
\centering
\scalebox{0.9}{ 
\begin{tabular}{l|ccccc}
\toprule
\multicolumn{1}{c|}{Method} & $\rightarrow$C & $\rightarrow$P & $\rightarrow$R & $\rightarrow$A & Avg. \\ 
\midrule
TPDS ($A$) & 59.1 & 81.7 & 81.7 & 71.6 & 73.5 \\  
LCFD-C-B32 ($B$) & 72.2 & 90.2 & 89.7 & 81.0 & 83.3 \\  
DIFO-C-B32 ($C$) & 70.4 & \textbf{90.8} & 88.8 & \textbf{82.3} & 83.1 \\  
\midrule
RCL ($A,B,C$) & \textbf{71.9} & 90.7 & \textbf{89.2} & 81.7 & \textbf{83.4} \\  
\bottomrule
\end{tabular}%
}
\label{tab:no_mllms}
\end{table}

\textbf{Synergy between RCL and MLLMs.} 
Table~\ref{tab:no_mllms} shows that without MLLMs, RCL provides only a marginal 0.1\% gain over the best existing SFDA method (LCFD-C-B32), indicating limited improvement when applied to standard adaptation approaches (TPDS, LCFD-C-B32, DIFO-C-B32). In contrast, integrating MLLMs (LLaVA-34B, ShareGPT4V-13B, InstBLIP-XXL) into RCL yields an 2.8\% improvement over the best MLLM model (LLaVa-34B) indicated in Table~\ref{tab:office-home} and a 6.4\% boost over RCL without MLLMs. Unlike ImageNet-pretrained models, MLLMs are trained on broad multimodal corpora, providing valuable auxiliary knowledge that enhances pseudo-label quality and adaptation effectiveness. These results confirm that traditional SFDA methods lack the generalization capacity of MLLMs, while RCL effectively leverages their complementary knowledge for improved adaptation.

\begin{figure}[!h]
\vspace{-1mm}
    \centering
    \includegraphics[width=0.5\textwidth]{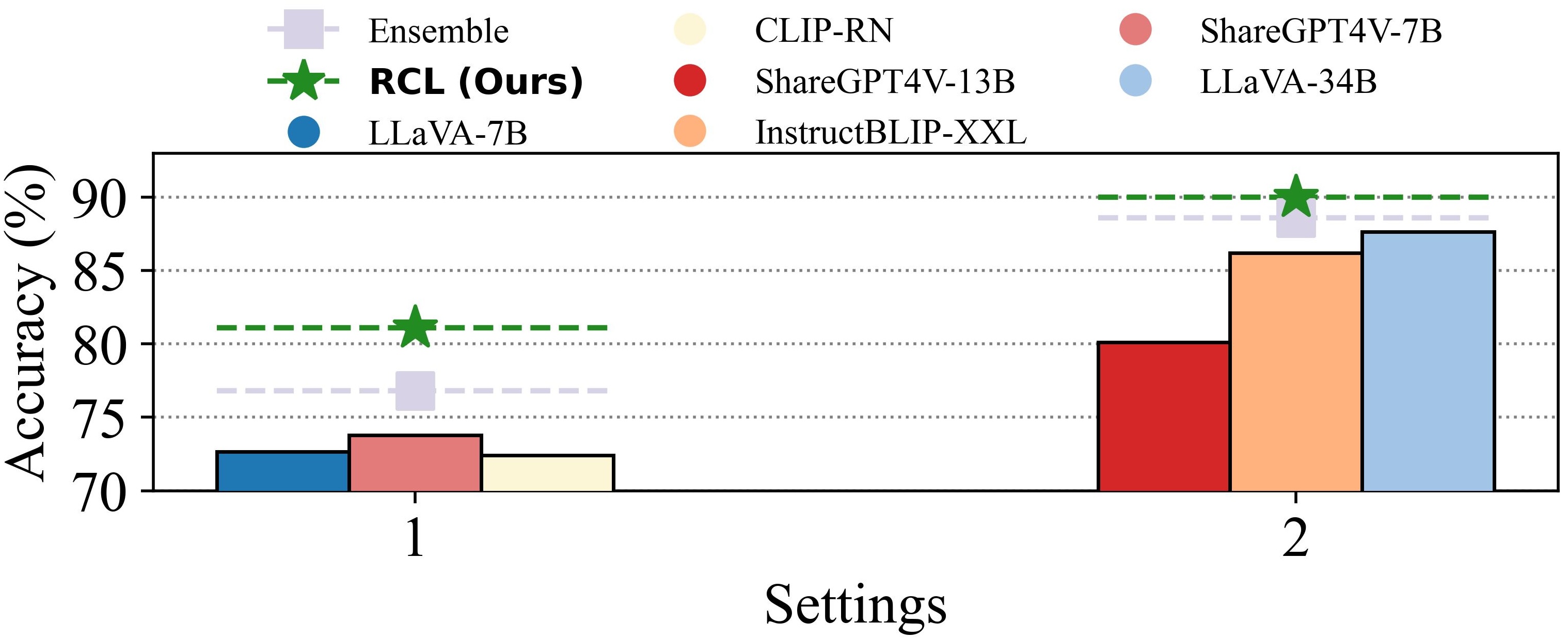}
    \vspace{-0.6cm}
    \caption{\small{RCL's sensitivity using weaker teacher MLLMs.}}
    \label{fig:mllm_sensitivity}
\end{figure}

\textbf{Sensitivity to the capability of MLLMs.}
Fig.~\ref{fig:mllm_sensitivity} compares two settings: (1) weaker MLLMs with lower zero-shot performance and (2) stronger MLLMs with higher ensemble accuracy. Labels are determined by majority vote, with ties assigned randomly. \emph{RCL consistently outperforms individual MLLMs and even their ensemble}, with the largest gap in Setting 1 (+4.3\%), where weaker MLLMs struggle. To understand whether RCL’s improvements stem from simple ensemble effects, we also considered a baseline that uses majority voting or averaged predictions from the frozen MLLMs directly. While these provide moderate gains over individual models, they lack consistency and fail to generalize as well as our distillation-based approach. This highlights the value of structured, curriculum-driven training over raw ensembling. Even in Setting 2, where MLLMs already perform well, RCL achieves a further +1.4\% improvement. \emph{These results demonstrate that RCL does not rely on strongest teacher models to be effective. While better MLLMs help, our method delivers large improvements even when only weak teacher MLLMs are available.}

\begin{figure}[bp]
\vspace{-1mm}
    \centering
    \includegraphics[width=0.45\textwidth]{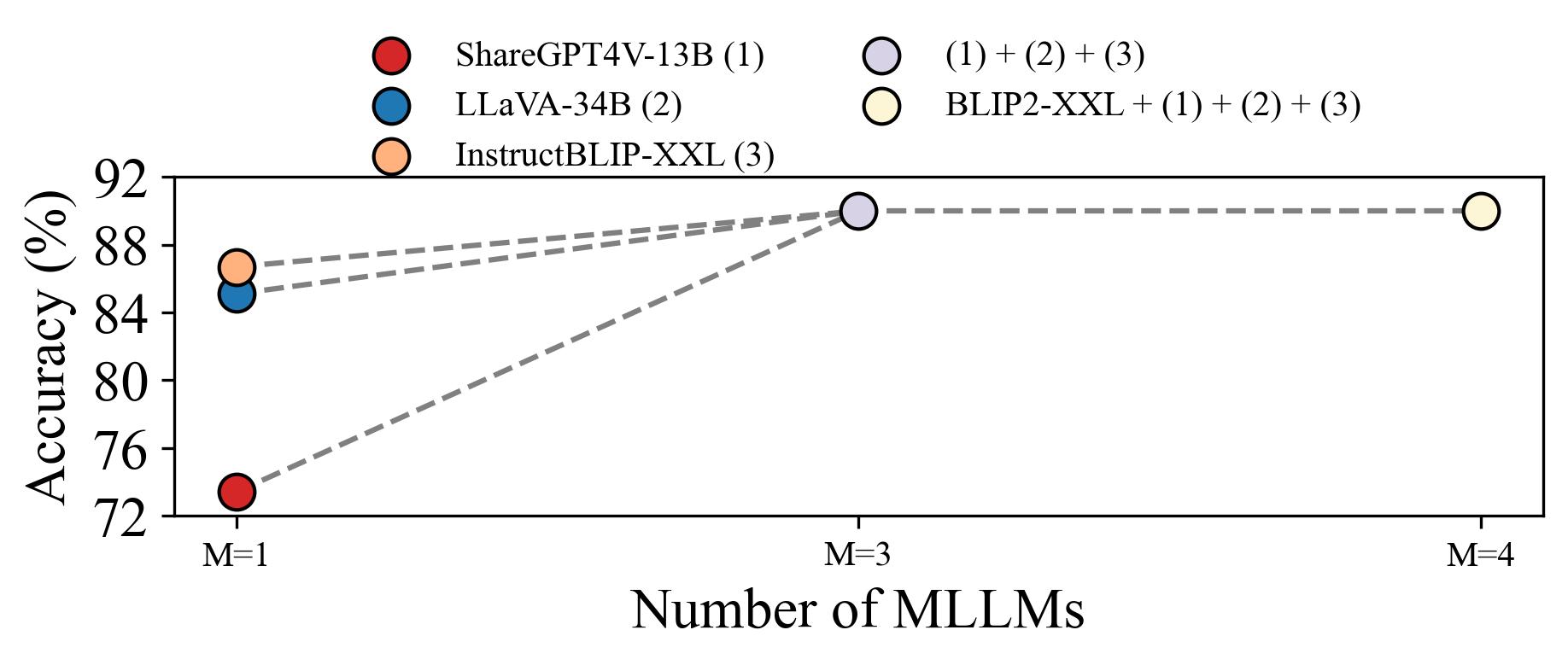}
    \vspace{-0.4cm}
    \caption{\small{Impact of number of MLLMs on RCL performance, the performance of using single MLLM is with RKT only.}}
    \label{fig:mllm_number}
\vspace{-3mm}
\end{figure}

\begin{figure}[tbp]
\vspace{-1mm}
    \centering
    \includegraphics[width=0.4\textwidth]{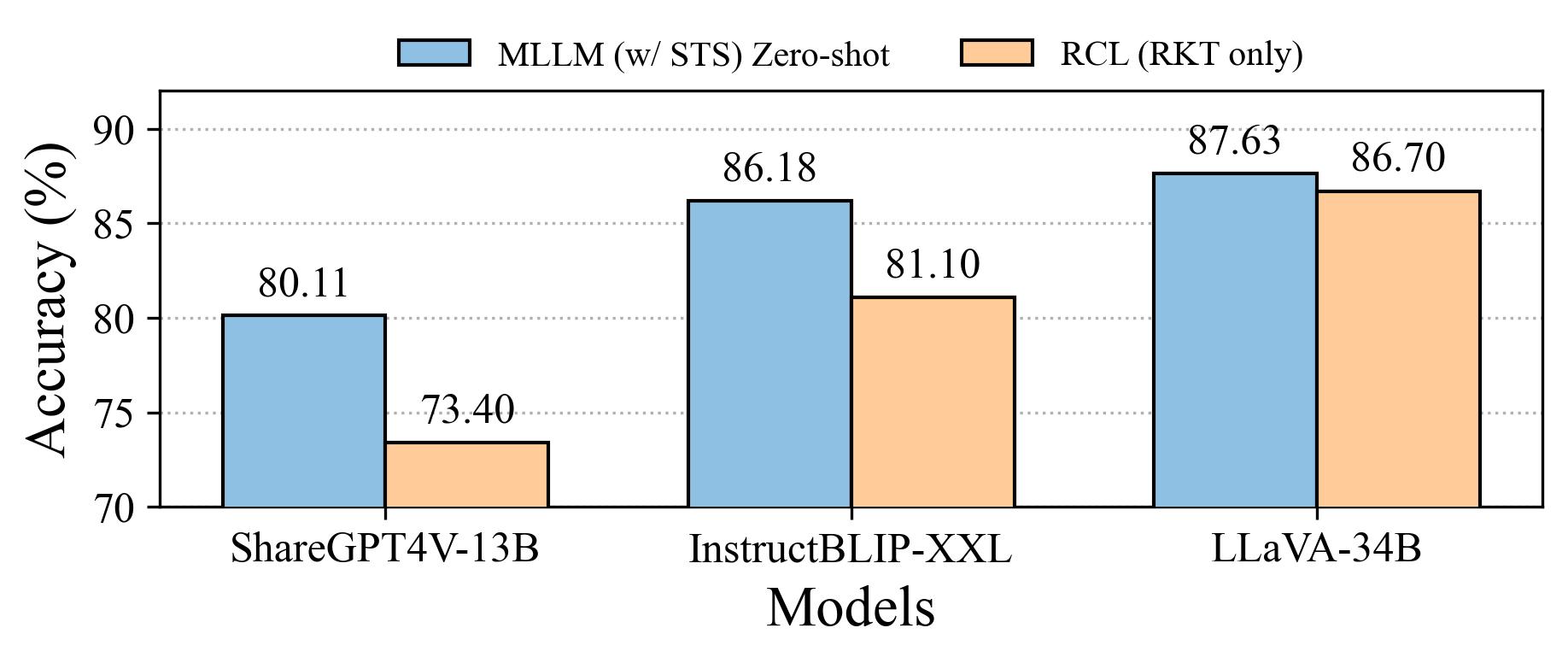}
    \vspace{-0.3cm}
    \caption{\small{Distillation from single MLLM to RCL cannot surpass the teacher model}}
    \label{fig:mllm_one_model}
\vspace{-4mm}
\end{figure}

\textbf{Number of MLLMs.}
Fig.~\ref{fig:mllm_number} shows the effect of using one, three, or four MLLMs. We use our strongest teacher MLLMs (ShareGPT-13B, InstructBLIP-XXL, LLaVA-34B) individually and combined, with BLIP2-XXL added for the 4-model ensemble. RCL gets the lowest accuracy using a single MLLM (Fig.~\ref{fig:mllm_one_model}), as RKT-only learning is bounded by its teacher performance, whereas it surpasses each MLLM by integrating and adapting their knowledge. RCL accuracy but saturates beyond three teacher MLLMs, while inference cost grows. Thus using three teachers offers a good trade-off.

\begin{table}[!h]
\caption{Ablation study on the choice of backbone (BB).}
\centering
\vspace{-3mm}
\scalebox{0.9}{ 
\begin{tabular}{c|c|cccc|c}
\toprule
\multirow{2}{*}{Method} & \multirow{2}{*}{BB} & \multicolumn{5}{c}{Office-Home} \\
 & & $\rightarrow$A & $\rightarrow$C & $\rightarrow$P & $\rightarrow$R & \multicolumn{1}{|c}{Avg.} \\ \midrule
DIFO-C-RN & RN50 & 79.3 & 63.1 & 87.7 & 87.5 & 79.4 \\
DIFO-C-B32 & RN50 & 82.3 & 70.4 & 90.8 & 88.3 & 83.1 \\
RCL (Ours) & RN18 & 89.1 & 81.5 & 95.1 & 92.6 & 89.6 \\
RCL (Ours) & RN50 & \textbf{89.3} & \textbf{82.3} & \textbf{95.3} & \textbf{92.9} & \textbf{90.0} \\
\bottomrule
\end{tabular}%
}
\label{tab:backbone_ablation}
\end{table}

\textbf{Knowledge transfer to a smaller backbone.} 
We investigate transferring to a smaller backbone for the scalability of SFDA. As shown in Table~\ref{tab:backbone_ablation}, RCL achieves similar performance with ResNet18 as with ResNet50, \emph{maintaining a +6.4\% accuracy advantage over state-of-the-art methods}. With twice the speed (2 GFLOPS v.s. 4 GFLOPS), \emph{RCL effectively distills pre-trained knowledge to a smaller backbone}, making it ideal for large-scale inference across diverse tasks. This demonstrates RCL's capability to efficiently leverage MLLM knowledge and adapt it to a lightweight model suitable for real-world deployment. The similar ResNet/ViT results suggest that reliability-gated multi-teacher distillation in RCL reduces backbone sensitivity.

\subsection{Discussion}

\textbf{Why MLLMs as teachers?}  
While RCL is teacher-agnostic, MLLMs are uniquely compelling frozen teachers for SFDA because they pair strong visual encoders with instruction-tuned LLMs, yielding richer semantics and task following than contrastive VLMs like CLIP. Large-scale instruction tuning has produced general-purpose MLLMs with strong zero-shot performance across diverse image understanding tasks~\cite{meng2024mmiu, wang2024muirbench, li2024seed, chen2024mllm, yue2024mmmu}, and domain-specific models further demonstrate their applicability in sensitive areas like medical imaging~\cite{li2023llava-med, li2025llava, lee2025cxr}. Specialized benchmarks~\cite{lozano2024micro, lai2025can, sepehri2024mediconfusion, yue2025medsg} highlight both zero-shot generalization and residual hallucinations, which our reliability-based multi-teacher distillation mitigates. Recent studies also show that MLLMs have caught up with, and in many cases surpassed, CLIP on standard image classification~\cite{liu2024revisiting}, which emphasizes their potential and promise for SFDA. We stress that RCL is not contingent on largest MLLMs, weaker MLLM/CLIP teachers remain competitive (see Fig.~\ref{fig:mllm_sensitivity}, Table~\ref{tab:no_mllms}). However, zero-shot MLLMs with STS already achieve strong results. We therefore treat MLLMs as a high-value option (richer supervision) rather than a requirement.

\textbf{Practicality.}  
MLLMs offer valuable foundation knowledge and broad generalization, but their large inference cost makes direct use impractical (see Table~\ref{tab:latency_comparison} in \textit{Supplementary}). In RCL, they are queried only once on the target dataset to produce pseudo-labels, which are cached for training the lightweight student; no prompt tuning, augmentations, or repeated inference is required. The main limitation is the one-time cost of running large MLLMs, but this can be reduced by substituting smaller MLLMs or CLIP/ViT teachers, which still yield competitive results (see Fig.~\ref{fig:mllm_sensitivity}, \textit{Supplementary} Table~\ref{tab:mllm_sensitivity}). Thus, RCL combines the semantic strength of foundation models with the efficiency of compact students, enabling practical deployment in real-world SFDA settings.

\section{Conclusion}

We introduce Reliability-based Curriculum Learning (RCL), a multi-teacher distillation framework that improves SFDA by progressively distilling from pseudo-labels ordered by teacher reliability. In this framework, MLLMs act as strong teachers and our proposed STS maps their open-ended, instruction-following outputs to class labels, enabling lightweight students that outperform both state-of-the-art SFDA methods and large MLLM baselines. Beyond SFDA, the same reliability-driven multi-teacher principles extend to broader adaptation and transfer settings in vision.

 \textbf{\textit{Limitations:}} RCL can inherit biases from pre-trained MLLMs, though consensus reduces dependence on any single teacher. Pseudo-labeling with large MLLMs is costly, but this one-time step can be amortized across domains given their general-purpose knowledge and growing adoption.

\section*{Acknowledgement}
This work is supported in part by UC Noyce Initiative and Child Family Endowed Professorship.

{
    \small
    \bibliographystyle{ieeenat_fullname}
    \bibliography{main}
}
\clearpage
\setcounter{page}{1}
\maketitlesupplementary

\noindent This supplementary material provides additional details and results for RCL. Sec.~\ref{appendix:details} includes training details and hyperparameters. Sec.~\ref{sec:appendix_all_results} discusses extended results, including ablations, latency and backbone analyses, and sensitivity to the number and quality of MLLM teachers. Secs.~\ref{appendix:mmr}–\ref{appendix:mllm} detail the MMR algorithm and MLLM pseudo-labeling pipeline, including prompt templates, STS alignment, and reliability distributions. Sec.~\ref{sec:extended_related_work} extends related works to further position RCL within SFDA/DA literature. Finally, Sec.~\ref{sec:extended_discussion} extends our discussions on RCL's choice for MLLMs and practicality as a general multi-teacher framework.

\section{Training Details}
\label{appendix:details}

The parameters used in the training process of RCL are shown in Table~\ref{tab:hyperparameters}. We use the Adam optimizer~\cite{kingma2015adam} for all experiments, conducted in PyTorch on NVIDIA A100 GPUs due to availability. However, A100s are not required for training. RCL only finetunes lightweight backbones such as ResNet-50, ResNet-18, and ViT-B/32. Importantly, MLLMs are used solely for zero-shot inference to generate pseudo-labels, which are cached once before adaptation. Unlike other VLM-based methods that require repeated inference or prompt tuning during training, RCL is efficient and requires no tuning or retraining of the MLLMs.

\begin{table}[h]
\centering
\caption{\label{tab:hyperparameters} Parameters for different datasets and methods}
\vspace{-0.2cm}
\setlength\tabcolsep{4pt}
\scalebox{0.68}{ 
\begin{tabular}{ccccccccccc} 
\toprule
& \multicolumn{3}{c}{Office-Home} & \multicolumn{3}{c}{DomainNet} & \multicolumn{3}{c}{VisDA} \\
\cmidrule(lr){2-4} \cmidrule(lr){5-7} \cmidrule(lr){8-10}
& RKT & SMKE & MMR & RKT & SMKE & MMR & RKT & SMKE & MMR \\
\midrule
learning rate & 1e-04 & 1e-05 & 1e-05 & 1e-05 & 1e-05 & 1e-05 & 1e-05 & 1e-06 & 1e-06 \\
$\tau$ & $-$ & 0.7 & 0.95 & $-$ & 0.7 & 0.9 & $-$ & 0.7 & 0.6 \\
batch size & 64 & 256 & 128 & 64 & 256 & 64 & 64 & 256 & 256 \\
max iter & 3000 & 5000 & 5000 & 8000 & 10000 & 5000& 6000 & 6000 & 5000 \\
\bottomrule
\vspace{-0.8cm}
\end{tabular}
}
\end{table}

\section{Extended Results}
\label{sec:appendix_all_results}
\subsection{Main Results}

Tables~\ref{tab:officehome-full}, \ref{tab:domainnet-full}, and \ref{tab:visda-full} provide extended results for the Office-Home, DomainNet, and VisDA datasets, respectively. RCL achieves the highest average accuracy across all datasets. For Office-Home and DomainNet, RCL attains the highest performance across all 12 adaptation tasks compared to existing methods. For VisDA, RCL achieves top results in 9 out of 12 categories and ranks second in the remaining three. In most scenarios, the second-best performance is primarily achieved using either DIFO-C-B32~\cite{tang2023sourceMLLM} or PSAT-GDA~\cite{tang2023progressive}. As previously discussed in Section~\ref{subsec:main_results}, DIFO employs task-specific prompt learning and requires tuning of the CLIP encoder. Conversely, PSAT-GDA uses a transformer-based SFDA approach, specifically training transformers for source guidance and domain alignment. Our method, RCL, does not require tuning or training of any large VLMs or transformers, instead relying solely on the zero-shot inference capabilities of MLLMs. Throughout the results, RCL consistently achieves significant performance improvements over existing methods.

\begin{table*}[h]
\caption{Accuracy (\%) on the \textbf{Office-Home}. SF denotes source-free, and CP, ViT denote the method uses CLIP, and ViT backbone respectively. We highlight the best result and underline the second-best one. (*) represents pre-trained CLIP/MLLM zero-shot performance. PADCLIP/ADCLIP ViT results as reported in respective paper, using ViT-B/16.} \label{tab:officehome-full}
\setlength\tabcolsep{3pt}
\begin{center}
\scalebox{0.76}{  
    \begin{tabular}{l|ccc|cccccccccccc|c}
    \toprule
    Method & SF & CP & ViT & A$\rightarrow$C & A$\rightarrow$P & A$\rightarrow$R & C$\rightarrow$A & C$\rightarrow$P & C$\rightarrow$R & P$\rightarrow$A & P$\rightarrow$C & P$\rightarrow$R & R$\rightarrow$A & R$\rightarrow$C & R$\rightarrow$P & Avg.\\
    \midrule
    Source & - & \textcolor{Red}{\xmark} & \textcolor{Red}{\xmark} & 44.7 & 64.2 & 69.4 & 48.3 & 57.9 & 60.3 & 49.5 & 40.3 & 67.2 & 59.7 & 45.6 & 73.0 & 56.7 \\
    \midrule
    DAPL-RN~\cite{ge2023domain} & \textcolor{Red}{\xmark} & \textcolor{Green}{\cmark} & \textcolor{Red}{\xmark} & 54.1 & 84.3 & 84.8 & 74.4 & 83.7 & 85.0 & 74.5 & 54.6 & 84.8 & 75.2 & 54.7 & 83.8 & 74.5 \\
    PADCLIP-RN~\cite{lai2023padclip} & \textcolor{Red}{\xmark} & \textcolor{Green}{\cmark} & \textcolor{Red}{\xmark} & 57.5 & 84.0 & 83.8 & 77.8 & 85.5 & 84.7 & 76.3 & 59.2 & 85.4 & 78.1 & 60.2 & 86.7 & 76.6 \\
    PADCLIP-ViT~\cite{lai2023padclip} & \textcolor{Red}{\xmark} & \textcolor{Green}{\cmark} & \textcolor{Green}{\cmark} & 76.4 & 90.6 & 90.8 & 86.7 & 92.3 & 92.0 & 86.0 & 74.5 & 91.5 & 86.9 & 79.1 & 93.1 & 86.7 \\
    ADCLIP-RN~\cite{singha2023ad} & \textcolor{Red}{\xmark} & \textcolor{Green}{\cmark} & \textcolor{Red}{\xmark} & 55.4 & 85.2 & 85.6 & 76.1 & 85.8 & 86.2 & 76.7 & 56.1 & 85.4 & 76.8 & 56.1 & 85.5 & 75.9 \\
    ADCLIP-ViT~\cite{singha2023ad} & \textcolor{Red}{\xmark} & \textcolor{Green}{\cmark} & \textcolor{Green}{\cmark} & 70.9 & 92.5 & 92.1 & 85.4 & 92.4 & 92.5 & 86.7 & 74.3 & 93.0 & 86.9 & 72.6 & 93.8 & 86.1 \\
    \midrule
    SHOT~\cite{SHOT_2021} & \textcolor{Green}{\cmark} & \textcolor{Red}{\xmark} & \textcolor{Red}{\xmark} & 56.7 & 77.9 & 80.6 & 68.0 & 78.0 & 79.4 & 67.9 & 54.5 & 82.3 & 74.2 & 58.6 & 84.5 & 71.9 \\
    NRC~\cite{yang2021nrc} & \textcolor{Green}{\cmark} & \textcolor{Red}{\xmark} & \textcolor{Red}{\xmark} & 57.7 & 80.3 & 82.0 & 68.1 & 79.8 & 78.6 & 65.3 & 56.4 & 83.0 & 71.0 & 58.6 & 85.6 & 72.2 \\
    GKD~\cite{tang2021model} & \textcolor{Green}{\cmark} & \textcolor{Red}{\xmark} & \textcolor{Red}{\xmark} & 56.5 & 78.2 & 81.8 & 68.7 & 78.9 & 79.1 & 67.6 & 54.8 & 82.6 & 74.4 & 58.5 & 84.8 & 72.2 \\
    AaD~\cite{yang2022attracting} & \textcolor{Green}{\cmark} & \textcolor{Red}{\xmark} & \textcolor{Red}{\xmark} & 59.3 & 79.3 & 82.1 & 68.9 & 79.8 & 79.5 & 67.2 & 57.4 & 83.1 & 72.1 & 58.5 & 85.4 & 72.7 \\
    AdaCon~\cite{chen2022contrastive} & \textcolor{Green}{\cmark} & \textcolor{Red}{\xmark} & \textcolor{Red}{\xmark} & 47.2 & 75.1 & 75.5 & 60.7 & 73.3 & 73.2 & 60.2 & 45.2 & 76.6 & 65.6 & 48.3 & 79.1 & 65.0 \\
    CoWA~\cite{lee2022confidence} & \textcolor{Green}{\cmark} & \textcolor{Red}{\xmark} & \textcolor{Red}{\xmark} & 56.9 & 78.4 & 81.0 & 69.1 & 80.0 & 79.9 & 67.7 & 57.2 & 82.4 & 72.8 & 60.5 & 84.5 & 72.5 \\
    SCLM~\cite{tang2022semantic} & \textcolor{Green}{\cmark} & \textcolor{Red}{\xmark} & \textcolor{Red}{\xmark} & 58.2 & 80.3 & 81.5 & 69.3 & 79.0 & 80.7 & 69.0 & 56.8 & 82.7 & 74.7 & 60.6 & 85.0 & 73.0 \\
    ELR~\cite{yi2023source} & \textcolor{Green}{\cmark} & \textcolor{Red}{\xmark} & \textcolor{Red}{\xmark} & 58.4 & 78.7 & 81.5 & 69.2 & 79.5 & 79.3 & 66.3 & 58.0 & 82.6 & 73.4 & 59.8 & 85.1 & 72.6 \\
    PLUE~\cite{litrico_2023_CVPR} & \textcolor{Green}{\cmark} & \textcolor{Red}{\xmark} & \textcolor{Red}{\xmark} & 49.1 & 73.5 & 78.2 & 62.9 & 73.5 & 74.5 & 62.2 & 48.3 & 78.6 & 68.6 & 51.8 & 81.5 & 66.9 \\
    TPDS~\cite{tang2023source} & \textcolor{Green}{\cmark} & \textcolor{Red}{\xmark} & \textcolor{Red}{\xmark} & 59.3 & 80.3 & 82.1 & 70.6 & 79.4 & 80.9 & 69.8 & 56.8 & 82.1 & 74.5 & 61.2 & 85.3 & 73.5 \\
    C-SFDA~\cite{karim2023c} & \textcolor{Green}{\cmark} & \textcolor{Red}{\xmark} & \textcolor{Red}{\xmark} & 60.3 & 80.2 & 82.9 & 69.3 & 80.1 & 78.8 & 67.3 & 58.1 & 83.4 & 73.6 & 61.3 & 86.3 & 73.5 \\
    PSAT-GDA~\cite{tang2023progressive} & \textcolor{Green}{\cmark} & \textcolor{Red}{\xmark} & \textcolor{Green}{\cmark} & 73.1 & 88.1 & 89.2 & 82.1 & 88.8 & 88.9 & 83.0 & 72.0 & 89.6 & 83.3 & 73.7 & 91.3 & 83.6 \\
    \midrule
    LCFD-C-RN~\cite{tang2024unified} & \textcolor{Green}{\cmark} & \textcolor{Green}{\cmark} & \textcolor{Red}{\xmark} & 60.1 & 85.6 & 86.2 & 77.2 & 86.0 & 86.3 & 76.6 & 61.0 & 86.5 & 77.5 & 61.4 & 86.2 & 77.6 \\
    LCFD-C-B32~\cite{tang2024unified} & \textcolor{Green}{\cmark} & \textcolor{Green}{\cmark} & \textcolor{Green}{\cmark} & 72.3 & 89.8 & 89.9 & 81.1 & 90.3 & 89.5 & 80.1 & 71.5 & 89.8 & 81.8 & 72.7 & 90.4 & 83.3 \\
    DIFO-C-RN~\cite{tang2023sourceMLLM} & \textcolor{Green}{\cmark} & \textcolor{Green}{\cmark} & \textcolor{Red}{\xmark} & 62.6 & 87.5 & 87.1 & 79.5 & 87.9 & 87.4 & 78.3 & 63.4 & 88.1 & 80.0 & 63.3 & 87.7 & 79.4 \\
    DIFO-C-B32~\cite{tang2023sourceMLLM} & \textcolor{Green}{\cmark} & \textcolor{Green}{\cmark} & \textcolor{Green}{\cmark} & 70.6 & 90.6 & 88.8 & 82.5 & 90.6 & 88.8 & 80.9 & 70.1 & 88.9 & 83.4 & 70.5 & 91.2 & 83.1 \\
    \midrule
    CLIP-RN~\cite{CLIP_2021}* & - & \textcolor{Green}{\cmark} & \textcolor{Red}{\xmark} & 51.7 & 85.0 & 83.7 & 69.3 & 85.0 & 83.7 & 69.3 & 51.7 & 83.7 & 69.3 & 51.7 & 85.0 & 72.4 \\
    LLaVA-34B~\cite{liu2023llava}* & - & \textcolor{Green}{\cmark} & \textcolor{Green}{\cmark} & 78.3 & 93.7 & 89.5 & 87.0 & 93.7 & 89.5 & 87.0 & 78.3 & 89.5 & 87.0 & 78.3 & 93.7 & 87.2 \\
    InstBLIP-XXL~\cite{dai2024instructblip}* & - & \textcolor{Green}{\cmark} & \textcolor{Green}{\cmark} & 82.0 & 91.6 & 88.8 & 82.2 & 91.6 & 88.8 & 82.2 & 82.0 & 88.8 & 82.2 & 82.0 & 91.6 & 86.2 \\
    ShrGPT4V-13B~\cite{chen2023sharegpt4v}* & - & \textcolor{Green}{\textcolor{Green}{\cmark}} & \textcolor{Green}{\textcolor{Green}{\cmark}} & 66.7 & 85.8 & 84.8 & 83.2 & 85.8 & 84.8 & 83.2 & 66.7 & 84.8 & 83.2 & 66.7 & 85.8 & 80.1 \\
    \midrule
     \textbf{RCL (Ours)} & \textcolor{Green}{\cmark} & \textcolor{red}{\xmark} & \textcolor{red}{\xmark} & \underline{82.5} & \underline{95.3} & \textbf{93.3} & \underline{89.1} & \textbf{95.3} & \textbf{92.7} & \textbf{89.3} & \textbf{82.4} & \underline{92.8} & \underline{89.4} & \underline{82.1} & \underline{95.4} & \underline{90.0} \\
    \textbf{RCL-ViT (Ours)} & \textcolor{Green}{\cmark} & \textcolor{red}{\xmark} & \textcolor{Green}{\cmark} & \textbf{83.1} & \textbf{95.7} & \underline{93.1} & \textbf{89.2} & \textbf{95.3} & \underline{92.6} & \underline{89.2} & \underline{82.3} & \textbf{92.9} & \textbf{90.0} & \textbf{83.2} & \textbf{95.5} & \textbf{90.2} \\
    \bottomrule
    \end{tabular}}
\end{center}
\end{table*}

\begin{table*}[h]
\caption{Accuracy (\%) on the \textbf{DomainNet}.} \label{tab:domainnet-full}
\setlength\tabcolsep{3pt}
\begin{center}
\scalebox{0.76}{  
    \begin{tabular}{l|ccc|cccccccccccc|c}
    \toprule
    Method & SF & CP & ViT & C$\rightarrow$P & C$\rightarrow$R & C$\rightarrow$S & P$\rightarrow$C & P$\rightarrow$R & P$\rightarrow$S & R$\rightarrow$C & R$\rightarrow$P & R$\rightarrow$S & S$\rightarrow$C & S$\rightarrow$P & S$\rightarrow$R & Avg.\\
    \midrule
    Source & - & \textcolor{Red}{\xmark} & \textcolor{Red}{\xmark} & 42.6 & 53.7 & 51.9 & 52.9 & 66.7 & 51.6 & 49.1 & 56.8 & 43.9 & 60.9 & 48.6 & 53.2 & 52.7 \\
    \midrule
    DAPL-RN~\cite{ge2023domain} & \textcolor{Red}{\xmark} & \textcolor{Green}{\cmark} & \textcolor{Red}{\xmark} & 72.4 & 87.6 & 65.9 & 72.7 & 87.6 & 65.6 & 73.2 & 72.4 & 66.2 & 73.8 & 72.9 & 87.8 & 74.8 \\
    ADCLIP-RN~\cite{lai2023padclip} & \textcolor{Red}{\xmark} & \textcolor{Green}{\cmark} & \textcolor{Red}{\xmark} & 71.7 & 88.1 & 66.0 & 73.2 & 86.9 & 65.2 & 73.6 & 73.0 & 68.4 & 72.3 & 74.2 & 89.3 & 75.2 \\
    \midrule
    SHOT~\cite{SHOT_2021} & \textcolor{Green}{\cmark} & \textcolor{Red}{\xmark} & \textcolor{Red}{\xmark} & 63.5 & 78.2 & 59.5 & 67.9 & 81.3 & 61.7 & 67.7 & 67.6 & 57.8 & 70.2 & 64.0 & 78.0 & 68.1 \\
    NRC~\cite{yang2021nrc} & \textcolor{Green}{\cmark} & \textcolor{Red}{\xmark} & \textcolor{Red}{\xmark} & 62.6 & 77.1 & 58.3 & 62.9 & 81.3 & 60.7 & 64.7 & 69.4 & 58.7 & 69.4 & 65.8 & 78.7 & 67.5 \\
    GKD~\cite{tang2021model} & \textcolor{Green}{\cmark} & \textcolor{Red}{\xmark} & \textcolor{Red}{\xmark} & 61.4 & 77.4 & 60.3 & 69.6 & 81.4 & 63.2 & 68.3 & 68.4 & 59.5 & 71.5 & 65.2 & 77.6 & 68.7 \\
    AdaCon~\cite{chen2022contrastive} & \textcolor{Green}{\cmark} & \textcolor{Red}{\xmark} & \textcolor{Red}{\xmark} & 60.8 & 74.8 & 55.9 & 62.2 & 78.3 & 58.2 & 63.1 & 68.1 & 55.6 & 67.1 & 66.0 & 75.4 & 65.4 \\
    CoWA~\cite{lee2022confidence} & \textcolor{Green}{\cmark} & \textcolor{Red}{\xmark} & \textcolor{Red}{\xmark} & 64.6 & 80.6 & 60.6 & 66.2 & 79.8 & 60.8 & 69.0 & 67.2 & 60.0 & 69.0 & 65.8 & 79.9 & 68.6 \\
    PLUE~\cite{litrico_2023_CVPR} & \textcolor{Green}{\cmark} & \textcolor{Red}{\xmark} & \textcolor{Red}{\xmark} & 59.8 & 74.0 & 56.0 & 61.6 & 78.5 & 57.9 & 61.6 & 65.9 & 53.8 & 67.5 & 64.3 & 76.0 & 64.7 \\
    TPDS~\cite{tang2023source} & \textcolor{Green}{\cmark} & \textcolor{Red}{\xmark} & \textcolor{Red}{\xmark} & 62.9 & 77.1 & 59.8 & 65.6 & 79.0 & 61.5 & 66.4 & 67.0 & 58.2 & 68.6 & 64.3 & 75.3 & 67.1 \\
    \midrule
    LCFD-C-RN~\cite{tang2024unified} & \textcolor{Green}{\cmark} & \textcolor{Green}{\cmark} & \textcolor{Red}{\xmark} & 75.4 & 88.2 & 72.0 & 75.8 & 88.3 & 72.1 & 76.1 & 75.6 & 71.2 & 77.6 & 75.9 & 88.2 & 78.0 \\
    LCFD-C-B32~\cite{tang2024unified} & \textcolor{Green}{\cmark} & \textcolor{Green}{\cmark} & \textcolor{Green}{\cmark} & 77.2 & 88.0 & 75.2 & 78.8 & 88.2 & 75.8 & 79.1 & 77.8 & 74.9 & 79.9 & 77.4 & 88.0 & 80.0 \\
    DIFO-C-RN~\cite{tang2023sourceMLLM} & \textcolor{Green}{\cmark} & \textcolor{Green}{\cmark} & \textcolor{Red}{\xmark} & 73.8 & 89.0 & 69.4 & 74.0 & \underline{88.7} & 70.1 & 74.8 & 74.6 & 69.6 & 74.7 & 74.3 & 88.0 & 76.7 \\
    DIFO-C-B32~\cite{tang2023sourceMLLM} & \textcolor{Green}{\cmark} & \textcolor{Green}{\cmark} & \textcolor{Green}{\cmark} & 76.6 & 87.2 & 74.9 & 80.0 & 87.4 & 75.6 & 80.8 & 77.3 & 75.5 & 80.5 & 76.7 & 87.3 & 80.0 \\
    \midrule
    LLaVA-34B~\cite{liu2023llava}* & - & \textcolor{Green}{\cmark} & \textcolor{Green}{\cmark} & 84.4 & 91.0 & 83.7 & 85.5 & 91.0 & 83.7 & 85.5 & 84.4 & 83.7 & 85.5 & 84.4 & 91.0 & 86.1 \\
    InstBLIP-XXL~\cite{dai2024instructblip}* & - & \textcolor{Green}{\cmark} & \textcolor{Green}{\cmark} & 82.5 & 89.0 & 83.0 & 86.7 & 89.0 & 83.0 & 86.7 & 82.5 & 83.0 & 86.7 & 82.5 & 89.0 & 85.3 \\
    ShrGPT4V-13B~\cite{chen2023sharegpt4v}* & - & \textcolor{Green}{\cmark} & \textcolor{Green}{\cmark} & 79.7 & 87.9 & 79.2 & 79.9 & 87.9 & 79.2 & 79.9 & 79.7 & 79.2 & 79.9 & 79.7 & 87.9 & 81.7 \\
    \midrule
    \textbf{RCL (Ours)} & \textcolor{Green}{\cmark} & \textcolor{Red}{\xmark} & \textcolor{Red}{\xmark} &\underline{87.6} & \underline{92.8} & \underline{87.9} & \underline{89.2} & \underline{92.7} & \underline{87.8} & \underline{89.6} & \underline{87.7} & \underline{87.6} & \underline{89.4} & \underline{87.5} & \underline{92.7} & \underline{89.4} \\
    \textbf{RCL-ViT (Ours)} & \textcolor{Green}{\cmark} & \textcolor{red}{\xmark} & \textcolor{Green}{\cmark} & \textbf{88.1} & \textbf{93.3} & \textbf{88.0} & \textbf{89.7} & \textbf{93.3} & \textbf{88.0} & \textbf{89.7} & \textbf{88.0} & \textbf{87.8} & \textbf{89.7} & \textbf{88.1} & \textbf{93.3} & \textbf{89.7} \\
    \bottomrule
    \end{tabular}}
\end{center}
\vspace{-5mm}
\end{table*}

\begin{table*}[h]
\caption{ Accuracy (\%) on the \textbf{VisDA-C}.}\label{tab:visda-full}
\vspace{-5mm}
\setlength\tabcolsep{4pt}
\begin{center}
\scalebox{0.75}{
    \begin{tabular}{l|ccc|cccccccccccc|c}
    \toprule
    Method & SF & CP & ViT & plane & bcyle & bus & car & horse & knife & mcycl & person & plant & sktbrd & train & truck & Avg.\\
    \midrule
    Source & - & \textcolor{Red}{\xmark} & \textcolor{Red}{\xmark} & 60.4 & 22.5 & 44.8 & 73.4 & 60.6 & 3.28 & 81.3 & 22.1 & 62.2 & 24.8 & 83.7 & 4.81 & 45.3 \\
    \midrule
    DAPL-RN~\cite{ge2023domain} & \textcolor{Red}{\xmark} & \textcolor{Green}{\cmark} & \textcolor{Red}{\xmark} & 97.8 & 83.1 & 88.8 & 77.9 & 97.4 & 91.5 & 94.2 & 79.7 & 88.6 & 89.3 & 92.5 & 62.0 & 86.9 \\
    PADCLIP-RN~\cite{lai2023padclip} & \textcolor{Red}{\xmark} & \textcolor{Green}{\cmark} & \textcolor{Red}{\xmark} & 96.7 & 88.8 & 87.0 & 82.8 & 97.1 & 93.0 & 91.3 & 83.0 & 95.5 & 91.8 & 91.5 & 63.0 & 88.5 \\
    ADCLIP-RN~\cite{singha2023ad} & \textcolor{Red}{\xmark} & \textcolor{Green}{\cmark} & \textcolor{Red}{\xmark} & 98.1 & 83.6 & 91.2 & 76.6 & 98.1 & 93.4 & 96.0 & 81.4 & 86.4 & 91.5 & 92.1 & 64.2 & 87.7 \\
    \midrule
    SHOT~\cite{SHOT_2021} & \textcolor{Green}{\cmark} & \textcolor{Red}{\xmark} & \textcolor{Red}{\xmark} & 95.0 & 87.4 & 80.9 & 57.6 & 93.9 & 94.1 & 79.4 & 80.4 & 90.9 & 89.8 & 85.8 & 57.5 & 82.7 \\
    NRC~\cite{yang2021nrc} & \textcolor{Green}{\cmark} & \textcolor{Red}{\xmark} & \textcolor{Red}{\xmark} & 96.8 & 91.3 & 82.4 & 62.4 & 96.2 & 95.9 & 86.1 & 90.7 & 94.8 & 94.1 & 90.4 & 59.7 & 85.9 \\
    GKD~\cite{tang2021model} & \textcolor{Green}{\cmark} & \textcolor{Red}{\xmark} & \textcolor{Red}{\xmark} & 95.3 & 87.6 & 81.7 & 58.1 & 93.9 & 94.0 & 80.0 & 80.0 & 91.2 & 91.0 & 86.9 & 56.1 & 83.0 \\
    AaD~\cite{yang2022attracting} & \textcolor{Green}{\cmark} & \textcolor{Red}{\xmark} & \textcolor{Red}{\xmark} & 97.4 & 90.5 & 80.8 & 76.2 & 97.3 & 96.1 & 89.8 & 82.9 & 95.5 & 93.0 & 92.0 & 64.7 & 88.0 \\
    AdaCon~\cite{chen2022contrastive} & \textcolor{Green}{\cmark} & \textcolor{Red}{\xmark} & \textcolor{Red}{\xmark} & 97.0 & 84.7 & 84.0 & 77.3 & 96.7 & 93.8 & 91.9 & 84.8 & 94.3 & 93.1 & 94.1 & 49.7 & 86.8 \\
    CoWA~\cite{lee2022confidence} & \textcolor{Green}{\cmark} & \textcolor{Red}{\xmark} & \textcolor{Red}{\xmark} & 96.2 & 89.7 & 83.9 & 73.8 & 96.4 & \textbf{97.4} & 89.3 & \textbf{86.8} & 94.6 & 92.1 & 88.7 & 53.8 & 86.9 \\
    SCLM~\cite{tang2022semantic} & \textcolor{Green}{\cmark} & \textcolor{Red}{\xmark} & \textcolor{Red}{\xmark} & 97.1 & 90.7 & 85.6 & 62.0 & 97.3 & 94.6 & 81.8 & 84.3 & 93.6 & 92.8 & 88.0 & 55.9 & 85.3 \\
    ELR~\cite{yi2023source} & \textcolor{Green}{\cmark} & \textcolor{Red}{\xmark} & \textcolor{Red}{\xmark} & 97.1 & 89.7 & 82.7 & 62.0 & 96.2 & 97.0 & 87.6 & 81.2 & 93.7 & 94.1 & 90.2 & 58.6 & 85.8 \\
    PLUE~\cite{litrico_2023_CVPR} & \textcolor{Green}{\cmark} & \textcolor{Red}{\xmark} & \textcolor{Red}{\xmark} & 94.4 & 91.7 & 89.0 & 70.5 & 96.6 & 94.9 & 92.2 & 88.8 & 92.9 & 95.3 & 91.4 & 61.6 & 88.3 \\
    TPDS~\cite{tang2023source} & \textcolor{Green}{\cmark} & \textcolor{Red}{\xmark} & \textcolor{Red}{\xmark} & 97.6 & 91.5 & 89.7 & 83.4 & 97.5 & 96.3 & 92.2 & 82.4 & \underline{96.0} & 94.1 & 90.9 & 40.4 & 87.6 \\
    C-SFDA~\cite{karim2023c} & \textcolor{Green}{\cmark} & \textcolor{Red}{\xmark} & \textcolor{Red}{\xmark} & 97.6 & 88.8 & 86.1 & 72.2 & 97.2 & 94.4 & 92.1 & 84.7 & 93.0 & 90.7 & 93.1 & 63.5 & 87.8 \\
    PSAT-GDA~\cite{tang2023progressive} & \textcolor{Green}{\cmark} & \textcolor{Red}{\xmark} & \textcolor{Green}{\cmark} & 97.5 & \underline{92.4} & 89.9 & 72.5 & \underline{98.2} & 96.5 & 89.3 & 55.6 & 95.7 & \textbf{98.2} & \underline{95.3} & 54.8 & 86.3 \\
    \midrule
    DIFO-C-RN~\cite{tang2023sourceMLLM} & \textcolor{Green}{\cmark} & \textcolor{Green}{\cmark} & \textcolor{Red}{\xmark} & \underline{97.7} & 87.6 & 90.5 & 83.6 & 96.7 & 95.8 & \underline{94.8} & 74.1 & 92.4 & 93.8 & 92.9 & 65.5 & 88.8 \\
    DIFO-C-B32~\cite{tang2023sourceMLLM} & \textcolor{Green}{\cmark} & \textcolor{Green}{\cmark} & \textcolor{Green}{\cmark} & 97.5 & 89.0 & \underline{90.8} & \underline{83.5} & 97.8 & 97.3 & 93.2 & 83.5 & 95.2 & 96.8 & 93.7 & \underline{65.9} & \underline{90.3} \\
    \midrule
    LLaVA-34B~\cite{liu2023llava}*~ & - & \textcolor{Green}{\cmark} & \textcolor{Green}{\cmark} & 99.4 & 97.3 & 94.8 & 83.9 & 98.9 & 95.8 & 95.9 & 80.9 & 92.7 & 98.8 & 97.4 & 68.9 & 92.1 \\
    InstBLIP-XXL~\cite{dai2024instructblip}*~ & - & \textcolor{Green}{\cmark} & \textcolor{Green}{\cmark} & 99.2 & 89.6 & 82.0 & 69.8 & 97.9 & 91.0 & 97.5 & 84.3 & 73.6 & 99.3 & 96.7 & 60.0 & 86.7 \\
    ShrGPT4V-13B~\cite{chen2023sharegpt4v}* & - & \textcolor{Green}{\cmark} & \textcolor{Green}{\cmark} & 99.2 & 94.7 & 90.8 & 87.9 & 98.3 & 92.1 & 97.3 & 68.0 & 96.3 & 95.6 & 96.8 & 68.2 & 90.4 \\
    \midrule
    \textbf{RCL (Ours)} & \textcolor{Green}{\cmark} & \textcolor{Red}{\xmark} & \textcolor{Red}{\xmark} & \textbf{99.5} & \textbf{96.1} & \textbf{92.6} & \textbf{89.4} & \textbf{99.1} & \underline{97.1} & \textbf{97.0} & \underline{85.8} & \textbf{96.6} & \underline{98.1} & \textbf{97.3} & \textbf{70.0} & \textbf{93.2} \\
    \bottomrule
    \end{tabular}
}
\end{center}
\end{table*}


\subsection{Extended Ablation Studies}
\label{appendix:ablation}

We provide additional ablation results, including hyper-parameter sensitivity (Sec.~\ref{subsubsec:hyperparam_sensitivity}), inference latency for RCL and the MLLM teachers (Sec.~\ref{subsubsec:latency}), robustness to different MLLM/VLM teacher choices (Sec.~\ref{subsubsec:mllm_capability}), the impact of each RCL component (RKT, SMKE, MMR; Sec.~\ref{subsubsec:rcl_components}), and knowledge transfer to smaller backbones (Sec.~\ref{subsubsec:smaller_backbones}).

\subsubsection{Sensitivity of hyper-parameters}
\label{subsubsec:hyperparam_sensitivity}
\begin{figure}[!h]
    \vspace{-0.4cm}
    \centering
    \includegraphics[width=1\linewidth]{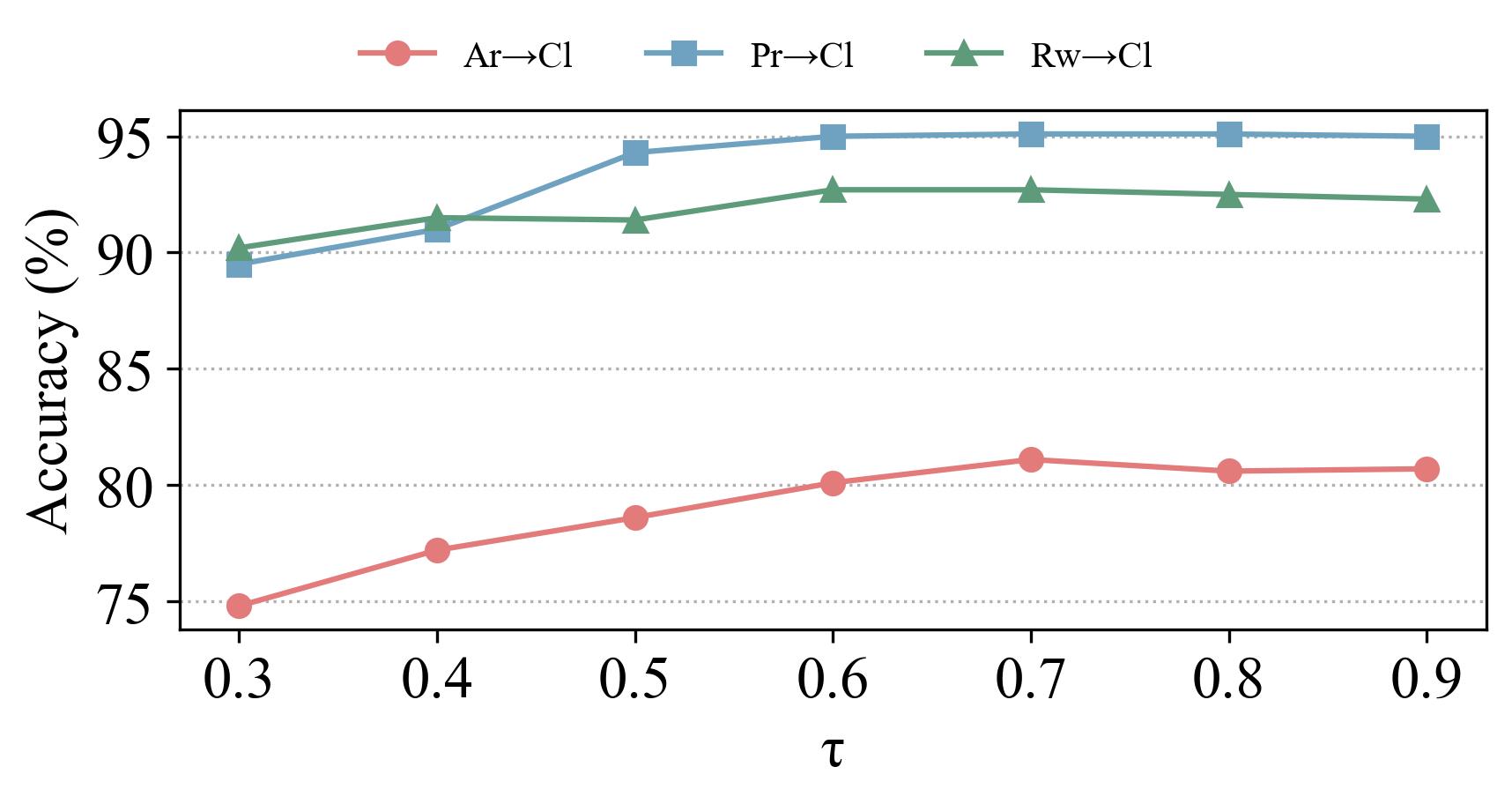}
    \vspace{-0.8cm}
    \caption{The effect of $\tau$ in SMKE.}
    \label{fig:tau}
\end{figure}

Fig.~\ref{fig:tau} shows the effect of the confidence threshold $\tau$ in SMKE, where $\tau$ values influence model performance by balancing self-correction with MLLM guidance. Higher $\tau$ values increase reliance on high-confidence pseudo-labels from the target model, while lower values depend more on MLLM-generated pseudo-labels. Fig.~\ref{fig:tau} highlights the optimal $\tau$ that enhances adaptation performance by effectively leveraging both reliable and less reliable pseudo-labels.

For the MMR, Table~\ref{tab:tau_accuracy_mmr} shows that $\tau$ = 0.95 provided a slight edge in accuracy, ensuring the model effectively utilizes its predictions while still considering MLLM guidance.

\begin{table}[h]
\centering
\caption{The effect of confidence threshold $\tau$ in MMR.}\label{tab:tau_accuracy_mmr}
\vspace{-3mm}
\scalebox{0.85}{ 
\begin{tabular}{c|cccccc}
\toprule
$\tau$ & 1.0 & 0.95 & 0.85 & 0.75 & 0.65 \\ \midrule
Ar$\rightarrow$Cl & 82.3 & 82.5 & 82.2 & 82.0 & 81.9 \\
Ar$\rightarrow$Pr & 95.4 & 95.3 & 95.3 & 95.3 & 95.3 \\
Ar$\rightarrow$Rw & 93.1 & 93.3 & 93.3 & 93.3 & 93.4 \\
\bottomrule
\end{tabular}
}
\end{table}

In Table~\ref{tab:lambda_target_accuracy}, we evaluate $\lambda$ values from 0 to 2.0. The results show that $\lambda$ = 0.5 consistently yielded the highest average accuracy across multiple domain adaptation tasks, suggesting it strikes a good balance between supervised learning and regularization.

\begin{table}[h]
\caption{Average accuracy for different $\lambda$ values in MMR.}\label{tab:lambda_target_accuracy}
\vspace{-5mm}
\begin{center}
\setlength\tabcolsep{5pt}
\scalebox{0.85}{ 
\begin{tabular}{c|c|c|c|c|c}
\toprule
$\lambda$ & $\rightarrow$C & $\rightarrow$P & $\rightarrow$R & $\rightarrow$A & Avg. \\ \midrule
0   & 82.0 & 95.3 & 92.8 & 89.1 & 89.8 \\
0.5 & 82.3 & 95.3 & 92.9 & 89.3 & \textbf{90.0} \\
1   & 82.3 & 95.3 & 92.8 & 89.2 & 89.9 \\
1.5 & 82.2 & 95.2 & 92.8 & 89.2 & 89.9 \\
2   & 82.3 & 95.2 & 92.8 & 89.1 & 89.9 \\
\bottomrule
\end{tabular}%
}
\end{center}
\end{table}

\subsubsection{Latency of deployed model}
\label{subsubsec:latency}
Table~\ref{tab:latency_comparison} illustrates the inference latency comparison of MLLMs and our proposed RCL. While LLaVA-34B, InstructBLIP-XXL, and ShareGPT4v-13B require latency over 1800ms per sample, RCL achieves a significantly faster inference speed of ~5ms per sample. When combining all three MLLMs for ensembling, over 7000ms latency is required. This over 378x improvement in computational efficiency validates our approach of distilling MLLM knowledge into a compact model rather than relying on direct MLLM inference, enabling practical deployment while achieving even better performance. 
All SFDA methods using ResNet-50 have comparable inference latency. RCL adds complexity only during initial pseudo-label generation with MLLMs but maintains efficient training and inference post-deployment while achieving superior performance. In contrast, DIFO~\cite{tang2023sourceMLLM} incorporates additional models (e.g., CLIP-RN, CLIP-ViT) in every training iteration, increasing computational overhead throughout training.

\begin{table}[h]
\centering
\small
\caption{Latency comparison for inferencing}
\begin{tabular}{@{}lcc@{}}
\toprule
\textbf{Model}       & \textbf{Avg Latency (ms / sample)} \\ \midrule
LLaVA-34B (w/STS)           & \(\sim \)2850   \\
ShrGPT4v-13B (w/STS)         & \(\sim \)1890   \\
InsBLIP-XXL (w/STS)         & \(\sim \)2740   \\
RCL (RN50)                 & \(\sim \)5      \\ \bottomrule
\end{tabular}
\label{tab:latency_comparison}
\end{table}

\begin{table*}[h]
\caption{\label{tab:mllm_sensitivity} RCL's sensitivity to MLLMs.} 
\vspace{-4mm}
\begin{center}
\setlength\tabcolsep{3pt} 
\scalebox{0.8}{ 
\begin{tabular}{cccccc|cccc|c}
\toprule
\multicolumn{6}{c|}{MLLMs/VLMs} & \multicolumn{5}{c}{Office-Home} \\
\multicolumn{1}{c}{LLaVA-34B} & \multicolumn{1}{c}{LLaVA-7B} & \multicolumn{1}{c}{InstBLIP-XXL} & \multicolumn{1}{c}{ShrGPT4V-13B} & \multicolumn{1}{c}{ShrGPT4V-7B} & \multicolumn{1}{c|}{CLIP-RN} & \multicolumn{1}{c}{$\rightarrow$A} & \multicolumn{1}{c}{$\rightarrow$C} & \multicolumn{1}{c}{$\rightarrow$P} & \multicolumn{1}{c}{$\rightarrow$R} & \multicolumn{1}{|c}{Avg.} \\ 
\midrule
 \textcolor{Gray}{\textcolor{Red}{\xmark}} & \textcolor{Gray}{\textcolor{Red}{\xmark}} & \textcolor{Gray}{\textcolor{Red}{\xmark}} & \textcolor{Green}{\textcolor{Green}{\cmark}} & \textcolor{Gray}{\textcolor{Red}{\xmark}} & \textcolor{Gray}{\textcolor{Red}{\xmark}} & 74.7 & 60.0 & 81.4 & 77.4 & 73.4 \\ 
 \textcolor{Gray}{\textcolor{Red}{\xmark}} & \textcolor{Gray}{\textcolor{Red}{\xmark}} & \textcolor{Green}{\textcolor{Green}{\cmark}} & \textcolor{Gray}{\textcolor{Red}{\xmark}} & \textcolor{Gray}{\textcolor{Red}{\xmark}} & \textcolor{Gray}{\textcolor{Red}{\xmark}} & 82.2 & 81.1 & 88.4 & 88.6 & 81.1 \\ 
 
 \textcolor{Green}{\textcolor{Green}{\cmark}} & \textcolor{Gray}{\textcolor{Red}{\xmark}} & \textcolor{Gray}{\textcolor{Red}{\xmark}} & \textcolor{Gray}{\textcolor{Red}{\xmark}} & \textcolor{Gray}{\textcolor{Red}{\xmark}} & \textcolor{Gray}{\textcolor{Red}{\xmark}} & 87.0 & 76.6 & 93.6 & 89.6 & 86.7 \\ 
 \midrule
 \textcolor{Gray}{\textcolor{Red}{\xmark}} & \textcolor{Green}{\textcolor{Green}{\cmark}} & \textcolor{Gray}{\textcolor{Red}{\xmark}} & \textcolor{Gray}{\textcolor{Red}{\xmark}} & \textcolor{Green}{\textcolor{Green}{\cmark}} & \textcolor{Green}{\textcolor{Green}{\cmark}} & 82.3 & 66.1 & 88.8 & 87.0 & 81.1 \\
 \textcolor{Gray}{\textcolor{Red}{\xmark}} & \textcolor{Gray}{\textcolor{Red}{\xmark}} & \textcolor{Green}{\textcolor{Green}{\cmark}} & \textcolor{Green}{\textcolor{Green}{\cmark}} & \textcolor{Gray}{\textcolor{Red}{\xmark}} & \textcolor{Green}{\textcolor{Green}{\cmark}} & 88.1 & 79.5 & 94.2 & \textbf{93.1} & 88.7 \\
 \textcolor{Green}{\textcolor{Green}{\cmark}} & \textcolor{Gray}{\textcolor{Red}{\xmark}} & \textcolor{Green}{\textcolor{Green}{\cmark}} & \textcolor{Green}{\textcolor{Green}{\cmark}} & \textcolor{Gray}{\textcolor{Red}{\xmark}} & \textcolor{Gray}{\textcolor{Red}{\xmark}} & \textbf{89.3} & \textbf{82.3} & \textbf{95.3} & 92.9 & \textbf{90.0} \\
 \bottomrule
\end{tabular}%
}
\end{center}
\end{table*} 

\begin{table*}[h]
\caption{\label{tab:mllm_consensus} Distribution (\%) of pseudo-labeled target samples by reliability on Office-Home: $D_R$ (all teachers agree), $D_{LR}$ (2/3 agree), and $D_{UR}$ (all disagree). Reported values show the proportion of each bin within the domain, alongside the resulting RCL accuracy. Total images per domain: A: 2,427, C: 4,365, P: 4,439, R: 4,357. Ensembles are the same as Table~\ref{tab:mllm_sensitivity}.} 
\vspace{-4mm}

\begin{center}
\setlength\tabcolsep{6pt} 
\scalebox{0.85}{ 
\begin{tabular}{l|c|ccc|c}
\toprule
MLLM/VLM Teacher Ensemble & Domain & $D_R$ & $D_{LR}$ & $D_{UR}$ & \textbf{Acc} \\
\midrule
LLaVA-7B + ShrGPT4V-7B + CLIP-RN & $\rightarrow$A & 53.7 & 34.4 & 11.9 & 82.3 \\
                                 & $\rightarrow$C & 35.1 & 43.4 & 21.7 & 66.1 \\
                                 & $\rightarrow$P & 64.5 & 27.3 & 8.2 & 88.8 \\
                                 & $\rightarrow$R & 63.9 & 28 & 8.1 & 87.0 \\
\midrule
InstBLIP-XXL + ShrGPT4V-13B + CLIP-RN & $\rightarrow$A & 60.3 & 29.5 & 10.2 & 88.1 \\
                                      & $\rightarrow$C & 41.1 & 35.6 & 25.5 & 79.5 \\
                                      & $\rightarrow$P & 71.4 & 22.0 & 6.6 & 94.2 \\
                                      & $\rightarrow$R & 70.6 & 22.8 & 6.6 & \textbf{93.1} \\
\midrule
LLaVA-34B + InstBLIP-XXL + ShrGPT4V-13B & $\rightarrow$A & 72.3 & 21.8 & 5.9 & \textbf{89.3} \\
                                        & $\rightarrow$C & 58.1 & 31.1 & 11.0 & \textbf{82.3} \\
                                        & $\rightarrow$P & 79.3 & 18.1 & 2.6 & \textbf{95.3} \\
                                        & $\rightarrow$R & 76.1 & 20.0 & 3.9 & 92.9 \\
\bottomrule
\end{tabular}%
}
\end{center}
\vspace{-0.3cm}
\end{table*}

\subsubsection{Sensitivity to the capability of MLLMs} 
\label{subsubsec:mllm_capability}
Table~\ref{tab:mllm_sensitivity} provides a detailed version of Figs.~\ref{fig:mllm_sensitivity} and~\ref{fig:mllm_one_model}. The upper part of the table shows the learning ability of the target model when learning from only one of the MLLMs, with the upper bound being the zero-shot performance of the individual MLLM. In contrast, RCL outperforms all the MLLMs it learns from, demonstrating that it effectively combines their knowledge and adapts it to the target domain through the curriculum learning process.
The lower part of Table~\ref{tab:mllm_sensitivity} illustrates the impact of replacing the best-performing MLLM (LLaVA-34B) with CLIP-RN50. Despite CLIP-RN50 having a 14.8\% lower accuracy on the target task and a 4.9\% lower average accuracy when combined with InstructBLIP-XXL and ShareGPT4V-13B, the performance of RCL only decreases by 1.3\%. Furthermore, even when using MLLM/CLIP models with an average accuracy of 72.9\%, RCL can still maintain an 81.1\% accuracy, comparable to SOTA methods. This demonstrates that RCL is robust to the choice of MLLMs, highlighting its effectiveness in leveraging knowledge from multiple sources and adapting to the target domain.

\subsubsection{Impact of RCL components} 
\label{subsubsec:rcl_components}
In Table~\ref{tab:rcl_ablation_full}, we present the full results on Office-Home to study the impact of RCL components. Across all adaptation tasks, RKT achieves the lowest performance, utilizing only reliable data. It is important to note that even with only RKT applied, RCL already reaches similar performance levels to existing methods. We further find that applying SMKE and MMR significantly boosts RCL performance across every task. This highlights the importance of our reliability scheme and the efficient utilization of all target data.

\begin{table*}[h]
\caption{Accuracy (\%) of various components of RCL training for select adaptation tasks.}\label{tab:rcl_ablation_full} 
\vspace{-5mm}
\begin{center}
\setlength\tabcolsep{3pt} 
\scalebox{0.8}{ 
\begin{tabular}{ccc|cccccccccccc|c}
\toprule
\multicolumn{3}{c|}{RCL} & \multicolumn{12}{c}{Office-Home} \\
\multicolumn{1}{c}{RKT} & \multicolumn{1}{c}{SMKE} & \multicolumn{1}{c|}{MMR} & \multicolumn{1}{c}{A$\rightarrow$C} & \multicolumn{1}{c}{A$\rightarrow$P} & \multicolumn{1}{c}{A$\rightarrow$R} & \multicolumn{1}{c}{C$\rightarrow$A} & \multicolumn{1}{c}{C$\rightarrow$P} & \multicolumn{1}{c}{C$\rightarrow$R} & \multicolumn{1}{c}{P$\rightarrow$A} & \multicolumn{1}{c}{P$\rightarrow$C} & \multicolumn{1}{c}{P$\rightarrow$R} & \multicolumn{1}{c}{R$\rightarrow$A} & \multicolumn{1}{c}{R$\rightarrow$C} & \multicolumn{1}{c}{R$\rightarrow$P} & \multicolumn{1}{|c}{Avg.} \\ \midrule
 \textcolor{Green}{\textcolor{Green}{\cmark}} & \textcolor{red}{\textcolor{Red}{\xmark}} & \textcolor{red}{\textcolor{Red}{\xmark}} & 73.5 & 89.3 & 88.2 & 82.6 & 89.1 & 87.9 & 82.7 & 73.2 & 88.2 & 83.1 & 73.2 & 89.4 & 83.3 \\
 \textcolor{Green}{\textcolor{Green}{\cmark}} & \textcolor{red}{\textcolor{Red}{\xmark}} & \textcolor{Green}{\textcolor{Green}{\cmark}} & 80.3 & 93.9 & 91.9 & 87.3 & 94.2 & 91.8 & 87.8 & 80.1 & 92.2 & 88.0 & 80.4 & 91.9 & 88.3  \\
 \textcolor{Green}{\textcolor{Green}{\cmark}} & \textcolor{Green}{\textcolor{Green}{\cmark}} & \textcolor{red}{\textcolor{Red}{\xmark}} & 81.1 & 95.1 & 92.7 & 88.5 & 95.0 & 92.3 & 88.5 & 80.8 & 92.4 & 88.6 & 80.8 & 95.3 & 89.3 \\
 \textcolor{Green}{\textcolor{Green}{\cmark}} & \textcolor{Green}{\textcolor{Green}{\cmark}} & \textcolor{Green}{\textcolor{Green}{\cmark}} & \textbf{82.5} & \textbf{95.3} & \textbf{93.3} & \textbf{89.1} & \textbf{95.3} & \textbf{92.7} & \textbf{89.3} & \textbf{82.4} & \textbf{92.8} & \textbf{89.4} & \textbf{82.1} & \textbf{95.4} & \textbf{90.0} \\ \bottomrule
\end{tabular}%
}
\end{center}
\end{table*}

\subsubsection{Knowledge transfer to a smaller backbone} 
\label{subsubsec:smaller_backbones}
Table~\ref{tab:backbone_full} provides the complete results on Office-Home for the choice of backbone architecture. We employ RCL with a ResNet18 backbone and find that, across all adaptation tasks, RCL with ResNet18 outperforms the existing state-of-the-art methods and performs comparably to RCL with a ResNet50 backbone. Our method is adept at efficiently leveraging knowledge from multiple MLLMs and using self-correction to distill it into a smaller architecture, which is vital for deployments with resource constraints.

\begin{table*}[h]
\caption{\label{tab:backbone_full} Impact of Backbone.}
\vspace{-5mm}
\begin{center}
\setlength\tabcolsep{3pt} 
\scalebox{0.8}{ 
\begin{tabular}{c|c|cccccccccccc|c}
\toprule
\multirow{2}{*}{Method} & \multirow{2}{*}{BB} & \multicolumn{12}{c}{Office-Home} \\
 & & \multicolumn{1}{c}{A$\rightarrow$C} & \multicolumn{1}{c}{A$\rightarrow$P} & \multicolumn{1}{c}{A$\rightarrow$R} & \multicolumn{1}{c}{C$\rightarrow$A} & \multicolumn{1}{c}{C$\rightarrow$P} & \multicolumn{1}{c}{C$\rightarrow$R} & \multicolumn{1}{c}{P$\rightarrow$A} & \multicolumn{1}{c}{P$\rightarrow$C} & \multicolumn{1}{c}{P$\rightarrow$R} & \multicolumn{1}{c}{R$\rightarrow$A} & \multicolumn{1}{c}{R$\rightarrow$C} & \multicolumn{1}{c}{R$\rightarrow$P} & \multicolumn{1}{|c}{Avg.} \\ \midrule
 DIFO-C-RN & RN50 & 62.6 & 87.5 & 87.1 & 79.5 & 87.9 & 87.4 & 78.3 & 63.4 & 88.1 & 80.0 & 63.3 & 87.7 & 79.4 \\
 DIFO-C-ViT & RN50 & 70.6 & 90.6 & 88.8 & 82.5 & 90.6 & 88.8 & 80.9 & 70.1 & 88.9 & 83.4 & 70.5 & 91.2 & 83.1 \\
 RCL (Ours) & RN18 & 81.2 & 95.3 & 92.8 & 88.9 & 94.9 & 92.4 & 88.8 & 81.7 & 92.4 & \textbf{89.5} & 81.6 & 95.1 & 89.6 \\
 RCL (Ours) & RN50 & \textbf{82.5} & \textbf{95.3} & \textbf{93.3} & \textbf{89.1} & \textbf{95.3} & \textbf{92.7} & \textbf{89.3} & \textbf{82.4} & \textbf{92.8} & 89.4 & \textbf{82.1} & \textbf{95.4} & \textbf{90.0} \\ \bottomrule
\end{tabular}%
}
\vspace{-5mm}
\end{center}
\end{table*}

\section{Multi-hot Masking Refinement Details}
\label{appendix:mmr}

Multi-hot Masking Refinement (MMR) (Sec.~\ref{sec:mmr}) refines pseudo-labels for unreliable samples via Multi-hot Masking and consistency loss, ensuring robust learning from all samples. Full pseudocode is provided in Alg.~\ref{alg:mmr}. 

\textbf{MMR Complexity and Effectiveness.} MMR demonstrates enhanced effectiveness for challenging adaptation tasks, with performance gains correlating to the proportion of unreliable samples. For example, the ($\rightarrow$ Clipart) achieves the most significant improvement of +1.4\% (80.9\% to 82.3\% in Table~\ref{tab:rcl_comp_ablation}), despite having the highest proportion of unreliable (R=0) samples (see Fig.~\ref{fig:class_dist_officehome}). These results indicate MMR's particular utility when dealing with large portions of unreliable labels, which aligns with real-world scenarios where MLLMs possess divergent knowledge spaces. MMR enables comprehensive data utilization through its ability to learn from all samples, rather than being limited to only high-confidence instances. Our lightweight inference model ensures deployment efficiency, achieving a practical latency of 5ms per sample.
Regarding the effectiveness of MMR components, we have conducted detailed ablation studies (Tables~\ref{tab:rcl_comp_ablation}, ~\ref{tab:lambda_target_accuracy}). These results demonstrate that both multi-hot masking and consistency loss contribute to the performance. Note that the MMR cannot be applied without Multi-hot Masking since the unreliable samples cannot be used by traditional pseudo-labeling methods. 

\begin{algorithm}[h]
\caption{Multi-hot Masking Refinement (MMR)}
\label{alg:mmr}
\begin{algorithmic}[1]
\STATE \textbf{Input:} Unlabeled dataset $\mathcal{D}_t$, confidence threshold $\tau$, model $f_{\theta_t}$, MLLM outputs $\hat{y}^{mi}$
\FOR{each sample $x_t^i \in \mathcal{D}_t$}
    \STATE Generate augmented views: $x_{t,weak}^i$ (weak) and $x_{t,strong}^i$ (strong)
    \STATE Obtain model predictions $\mathbf{z}_{t,weak}^i$ and $\mathbf{z}_{t,strong}^i$
    \STATE Compute confidence score $p_t^i = \operatorname*{max}_c {\mathbf{z}_{t,weak}^i}$
    \IF{$p_t^i \geq \tau$} 
        \STATE Assign pseudo-label $\tilde{y}^i = \operatorname*{argmax}_C \mathbf{z}_{t,weak}^i$
    \ELSE
        \STATE Compute multi-hot mask $\mathbf{m}^i$ from MLLM outputs:
        \[
        \mathbf{m}^i = 1 - \prod_{m=1}^M (1 - \mathbbm{1}(\hat{y}^{mi}))
        \]
        \STATE Refine pseudo-label:
        \[
        \tilde{y}^i =
        \begin{cases}
        \operatorname*{argmax}_C(\mathbf{z}_t^i), & \text{if } p_t^i \geq \tau, \\
        \operatorname*{argmax}_C(\mathbf{z}_t^i \odot \mathbf{m}^i), & \text{if } p_t^i < \tau,
        \end{cases}
        \]
    \ENDIF
    \STATE Compute supervised loss:
    \[
    \mathcal{L}_{\text{sup}} = -\tilde{y}^i \cdot \log f_{\theta_t}(x_{t,weak}^i)
    \]
    \STATE Compute consistency loss:
    \[
    \mathcal{L}_{\text{cons}} = \frac{1}{M} \sum_{m=1}^{M} \sum_{i=1}^{N_t}  \text{H}(\tilde{y}^i, \mathbf{z}_{st}^i)
    \]
    \STATE Update model by minimizing:
    \[
    \mathcal{L}_{\text{MMR}} = \mathcal{L}_{\text{sup}} + \lambda_{\text{cons}} \mathcal{L}_{\text{cons}}
    \]
\ENDFOR
\end{algorithmic}
\end{algorithm}

\section{MLLM Pseudo-labeling Details}
\label{appendix:mllm}

In this section, we provide details on prompt templates (Sec.~\ref{subsec:prompt_templates}), and discussions on pseudolabel distribution (Sec.~\ref{subsec:pseudolabel_dist}) and prompt sensitivity (Sec.~\ref{subsec:prompt_sensitivity}).

\subsection{Prompt Templates} 
\label{subsec:prompt_templates}

To facilitate pseudo-labeling, we design specific prompt templates for different MLLMs:

\textbf{LLaVA models:}
\textit{"Question: What is the closest name from this list to describe the object in the image? Return the name only. <class names>"}

\textbf{ShareGPT4V models:}
\textit{"Question: What is the closest name from this list to describe the object in the image? List: <class names> Return the closest name from the list only. Use *exact* names from the list only. Answer:"}

\textbf{InstructBLIP models:}
\textit{"Question: What is the closest name from this list to describe the object in the image? <class names>. Use the closest name from the list only. Pick the answer from the list only. Answer:"}

For the STS calculations, we utilize all-MiniLM-L6-v2 to generate vector representations of both the outputs from the MLLMs and the text options~\cite{reimers-2020-multilingual-sentence-bert}.

\subsection{Prompt Sensitivity}
\label{subsec:prompt_sensitivity}
To evaluate robustness of our STS strategy, we tested LLaVA-34B on OfficeHome under multiple prompt templates. Beyond the default template, we considered (i) a variant with added \emph{domain information}, (ii) a \emph{naive template} phrased as \textit{``What is this image from given list below \{class\_list\}?''}, (iii) a \emph{paraphrased class list}, and (iv) a list with injected \emph{distractor labels}.  
For \textit{paraphrased class names}, we manually curated synonyms or simplified forms of Office-Home categories (e.g., \textit{``alarm clock'' → ``clock''}, \textit{``computer monitor'' → ``monitor''}, \textit{``filing cabinet'' → ``cabinet''}, \textit{``push pin'' → ``thumbtack''}). While MLLM outputs differ since they pick from modified class lists, STS reliably maps predictions back to the original label set, showing consistency under lexical variation.  
For \textit{distractor labels}, we manually injected unrelated category names sampled from DomainNet and VisDA into the candidate list, simulating noisier prompts where irrelevant options appear alongside correct classes. These variations stress-test both STS alignment and consensus-driven reliability, showing that RCL remains stable despite prompt perturbations.

Results are shown in Table~\ref{tab:prompt_sensitivity}. Across all four Office-Home transfers, performance varies by less than 2\% absolute between prompt templates. While domain information slightly reduces accuracy, and paraphrases/distractors introduce minor drops, the overall averages remain stable.

\begin{table}[h]
\centering
\footnotesize
\caption{Prompt sensitivity analysis on Office-Home using STS. Zero-shot+STS accuracy with different prompt templates for LLaVA-34B show small fluctuations.}
\renewcommand{\arraystretch}{1.2}
\setlength{\tabcolsep}{5pt}
\begin{tabular}{lccc|c}
\toprule
\textbf{Prompt Template} & \textbf{Ar→Cl} & \textbf{Ar→Pr} & \textbf{Ar→Rw} & \textbf{Avg.} \\
\midrule
Naive prompt        & 76.80 & 92.10 & 87.45 & 85.71 \\
\textbf{Default prompt (D)}          & \textbf{78.35} & \textbf{93.78} & \textbf{89.58}  & \textbf{87.19} \\
D + Domain info     & 77.54 & 93.08 & 88.73 & 86.25 \\
D + Paraphrased classes   & 76.20 & 92.45 & 87.90 & 85.20 \\
D + Distractor labels     & 76.90 & 92.70 & 88.25 & 85.79 \\
\bottomrule
\end{tabular}
\label{tab:prompt_sensitivity}
\end{table}

\subsection{Pseudo-label Distributions} 
\label{subsec:pseudolabel_dist}
\begin{figure}[tbp]
    \centering
    \includegraphics[width=1\linewidth]{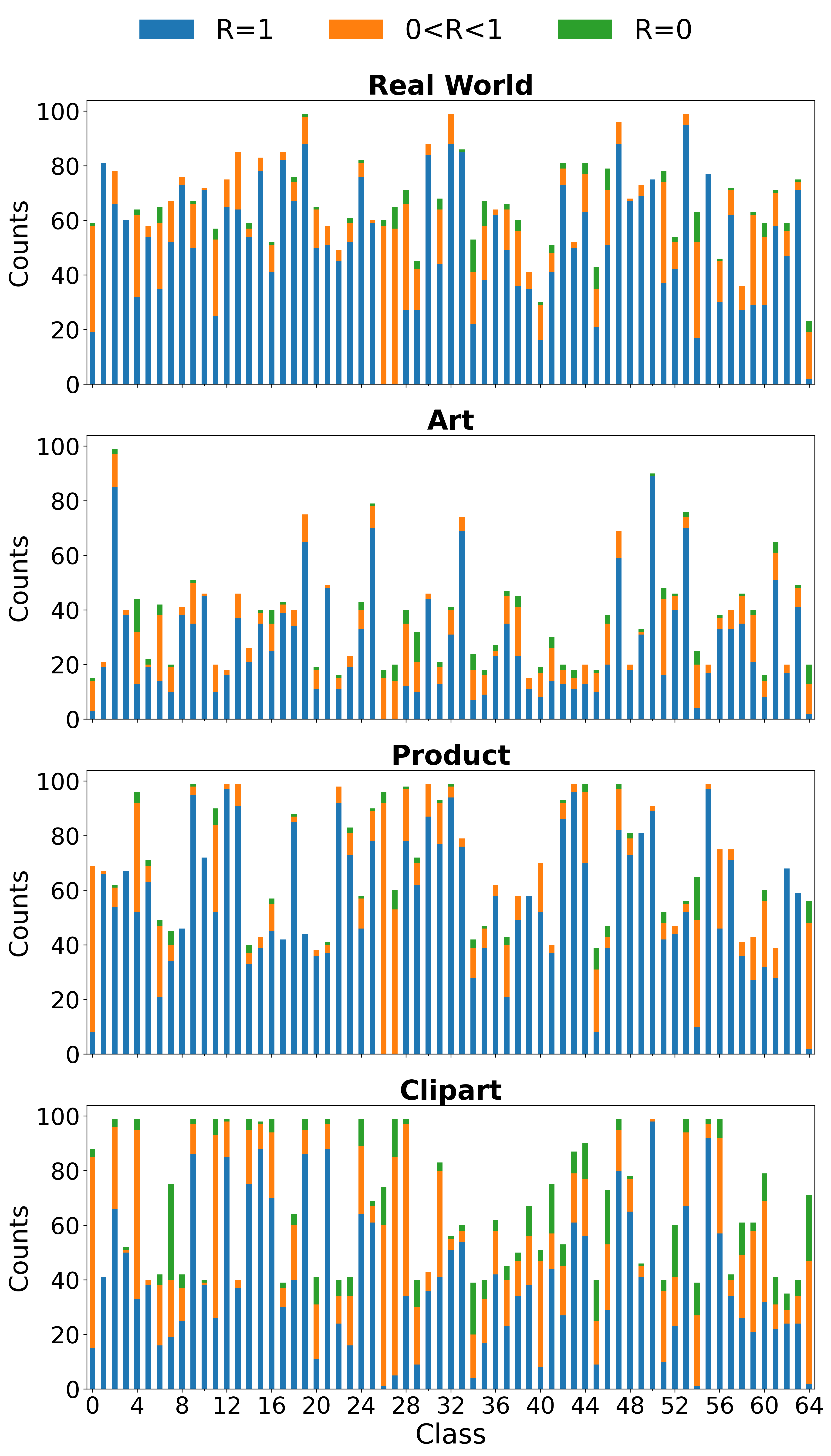}
    \caption{Per-class counts of pseudo-labels on Office-Home using LLaVA-34B, InstructBLIP-XXL, and ShareGPT4v-13B.}
    \label{fig:class_dist_officehome}
    \vspace{-4mm}
\end{figure}

Fig.~\ref{fig:class_dist_officehome} demonstrates the distribution of sample reliability across different classes within the Office-Home dataset. We observe that for various classes across domains, there are few samples with R=1 (high reliability). This emphasizes the importance of utilizing less reliable data, a fundamental aspect of our approach, which is integrated into our SMKE and MMR techniques. To highlight our point, if we observe Table~\ref{tab:rcl_comp_ablation}, it is evident that using only R=1 data (applying only RKT) results in notably lower accuracies for the Clipart domain (while adaptations to other domains achieve over 80\%, Clipart is lowest at 73.3\%). As Fig.~\ref{fig:class_dist_officehome} indicates, the Clipart domain has a higher proportion of pseudo-labels where 0<R<1. By employing SMKE and MMR, we manage to enhance performance in adapting to the Clipart domain by +9\% (to 82.3\%).
We also provide the distributions for $D_R$, $D_{LR}$, and $D_{UR}$ across various teacher ensembles and respective RCL accuracy in Table~\ref{tab:mllm_consensus}.

\section{Extended Related Work on SFDA and DA}
\label{sec:extended_related_work}

\textbf{SFDA Methods.}
Early SFDA methods adapted ImageNet-pretrained models using pseudo-label refinement, such as SHOT~\cite{SHOT_2021}, NRC~\cite{NRC_2022}, and G-SFDA~\cite{yang2021generalized}. These works emphasize clustering, entropy minimization, or distribution alignment under the no-source setting. More recent SFDA studies refine pseudo-labeling with neighborhood consistency~\cite{tang2021nearest, tang2022semantic} or target-structure guidance~\cite{chenself, litrico2023guiding}. TPDS~\cite{lai2024empowering} introduced target-prototype distribution smoothing for improved calibration but still relies on strong pre-trained features (e.g., CLIP). DIFO~\cite{tang2023sourceMLLM}, the current SOTA, adapts CLIP through prompt learning and distillation, progressively aligning CLIP with target data before transferring to a task-specific model. While effective, these works typically exploit a single teacher or finetune a large VLM, making them less scalable when source data is unavailable or when deploying lightweight students is required.

\textbf{Prompt-based DA with CLIP.}
Although not strictly SFDA, prompt-based DA methods highlight another direction. Chen et al.~\cite{chen2023multi} proposed multi-prompt alignment for multi-source DA; Phan et al.~\cite{hoang2024enhancing} improved cross-domain transfer by aligning prompt gradients; Ge et al.~\cite{ge2023domain} developed prompt learning for domain adaptation; and Vuong et al.~\cite{vuong2025preserving} preserved clusters in prompt learning for UDA. These approaches require source data or the ability to tune CLIP prompts, unlike SFDA where such access is not allowed. We include them here to show that our reliability-driven framework complements prompt-based DA by demonstrating how frozen teachers (MLLMs or CLIP) can be exploited without modification.

\textbf{Positioning of Our Work.}
Our RCL framework sits at the intersection of these lines: (1) like SFDA methods, it respects the no-source constraint, but unlike prior work, it incorporates multiple teachers; (2) like CLIP-based DA methods, it leverages large pretrained multimodal models, but without prompt tuning or finetuning; and (3) unlike DIFO or TPDS, it distills supervision into compact students by explicitly modeling teacher reliability and progressively incorporating all target data through a curriculum. This distinction is critical: RCL is not limited to a single CLIP teacher or large MLLMs, but provides a general strategy for multi-teacher SFDA under noisy pseudo-labels.

\section{Extended Discussions}
\label{sec:extended_discussion}

\subsection{Why use MLLMs?}

\textbf{MLLMs as Teachers.}  
A key motivation for adopting MLLMs is their role as foundation models that pair strong visual encoders with instruction-tuned LLMs, yielding broader semantics and instruction-following ability than contrastive VLMs like CLIP. Large-scale visual instruction tuning has produced general-purpose MLLMs with strong zero-shot performance across diverse image understanding tasks, as shown in multi-discipline benchmarks~\cite{meng2024mmiu, wang2024muirbench, li2024seed, chen2024mllm, yue2024mmmu}. Beyond general benchmarks, domain-specific MLLMs such as LLaVA-Med~\cite{li2023llava-med} and biomedical/industrial evaluations like MicroVQA~\cite{lozano2024micro}, and others~\cite{lai2025can, sepehri2024mediconfusion}, show their practical availability where task-specific models are scarce. Importantly, recent studies show that MLLMs have now caught up with and in many cases surpassed CLIP on standard image classification benchmarks~\cite{liu2024revisiting}, highlighting their promise as future-proof teachers for SFDA. RCL provides a recipe to distill such broad supervision into compact, task-specific students without prompt tuning, teacher training, multiple inferences per sample, or source data access. In our pipeline, noisy MLLM predictions or hallucinations are naturally handled as unreliable labels, mitigated through reliability-based multi-teacher distillation.

\textbf{Robustness to Teacher Variants.}  
At the same time, RCL is not contingent on large MLLMs. Our framework is fundamentally teacher-agnostic, focusing on multi-teacher curriculum learning under noisy pseudo-labels. Stress tests confirm this: replacing an MLLM with CLIP-RN50 reduces Office-Home accuracy by only 1.3\% (Supp. Table 10), while ensembles of weaker or smaller teachers (72.9\% average zero-shot accuracy) still achieve 81.1\% with RCL, surpassing strong CLIP-only SFDA baselines. Moreover, with modest student backbones (RN18/50), RCL outperforms ViT-based CLIP methods (Table 4, main), showing that the gains stem from reliability-driven distillation rather than teacher scale. Thus, MLLMs serve as a strong upper-bound case that demonstrates the potential of foundation models for SFDA, while RCL remains robust and effective with CLIP or heterogeneous teacher ensembles. We summarize key differences in using MLLM vs. CLIP/VLM models as techers in Table~\ref{tab:mllm_vs_clip_teachers}.

\subsection{Practicality of RCL.}  
Although MLLMs provide strong zero-shot classification, their direct use for classification is computationally expensive and unsuitable for large-scale or real-time deployment. RCL addresses this by requiring only a single inference pass of MLLMs (or other teachers) over the target dataset. These pseudo-labels are cached and used throughout training, meaning the heavy models need not be invoked again. In contrast, CLIP-based prompt learning often requires repeated forward passes per image across multiple prompts, and direct MLLM inference is prohibitive for deployment. By distilling cached pseudo-labels into a lightweight target model, RCL produces compact students that retain the benefits of foundation knowledge while being orders of magnitude cheaper to train and deploy. This design enables real-world SFDA usage: practitioners can harness the strengths of foundation models without maintaining them in production, instead deploying efficient students adapted with RCL.

\begin{table*}[h]
\centering
\footnotesize
\caption{Motivation and comparisons for MLLM vs. CLIP/VLM as teacher models.} \label{tab:mllm_vs_clip_teachers}
\renewcommand{\arraystretch}{1.15}
\setlength{\tabcolsep}{5pt}
\begin{tabular}{l|p{5.2cm}|p{5.2cm}}
\toprule
\textbf{Category} & \textbf{MLLM Teachers} & \textbf{CLIP / VLM Teachers} \\
\midrule
Efficacy & Richer semantics, reasoning, instruction-following. Zero-shot MLLMs w/ STS already surpass CLIP baselines (Tab.~1,2). RCL w/ MLLMs improves +6.4\% over CLIP-only RCL (Tab.~4). & Strong contrastive features, but limited to alignment. Require prompt tuning / finetuning for strong SFDA (e.g., DIFO). RCL with CLIP substitution still competitive (Supp. Tab.~10). \\
\midrule
Computation & High per-query latency (1800–2800ms, Supp. Tab.~9). In RCL, used \emph{once} for caching pseudo-labels; no repeated inference. & Lower latency per query, but prompt-learning methods require repeated inference per epoch, increasing total cost. \\
\midrule
Deployment & Not suitable for on-device deployment due to size. Distillation via RCL yields compact students ($\sim$5ms/sample). & More deployable, but adaptation pipelines (e.g., DIFO) still depend on CLIP during training, raising training-time overhead. \\
\midrule
Robustness & Aggregating diverse MLLMs mitigates hallucination; reliability scoring filters noise. Gains persist even with weaker MLLMs (Supp. Fig.~6, Tab.~10). & Ensembles can cover variance, but raw ensembling underperforms RCL distillation (Supp. Fig.~8). \\
\midrule
Generalizability & Encodes broad, multi-domain knowledge (general + domain-specific MLLMs). Strong upper bound for SFDA and beyond. & Strong on vision tasks but limited cross-domain reasoning. Less effective in medical/industrial applications where MLLM knowledge is valuable. \\
\bottomrule
\end{tabular}
\end{table*}

\subsection{RCL as a General Multi-Teacher SFDA Framework}

While we highlight MLLMs in the main paper, RCL can be seen as a general multi-teacher curriculum distillation framework for SFDA. The core components (RKT, SMKE, and MMR) are teacher-agnostic, working on pseudo-label reliability rather than specific architectures. Our proposed STS is what makes MLLMs viable teachers by converting free-form text outputs into closed-set predictions; for models like CLIP or ViTs, this conversion is not required and the same RCL can be applied.

We validate this generality through multiple experiments:
\begin{enumerate}
    \item CLIP substitution in Table~\ref{tab:mllm_sensitivity}. Replacing an MLLM with CLIP-RN50 reduces accuracy by only 1.3\%.
    \item Weaker/smaller teachers in Table~\ref{tab:mllm_sensitivity}. Even ensembles with an average of 72.9\% zero-shot accuracy achieve 81.1\% with RCL, surpassing strong CLIP-based SFDA methods.
    \item Comparison to ViT-based baselines (TPDS~\cite{tang2023source}, LCFD~\cite{tang2024unified}, DIFO~\cite{tang2023sourceMLLM}) in Table~\ref{tab:no_mllms} (main paper). RCL-trained ResNet student outperforms ViT-based CLIP prompt-learning methods, highlighting that the framework (not teacher scale) is the main driver.
\end{enumerate}

Thus, while MLLMs provide richer semantics and broad applicability (e.g., medical, industrial domains where task-specific teachers are unavailable), RCL is not confined to MLLMs. It is a robust, scalable framework that can integrate diverse teacher families, making it broadly useful for SFDA across settings. Future work can assess its viability and adoption across different tasks beyond SFDA.

\end{document}